\documentclass{article}

\PassOptionsToPackage{numbers, compress}{natbib}

\usepackage[main, final]{neurips_2026}

\usepackage[utf8]{inputenc}
\usepackage[T1]{fontenc}
\usepackage{microtype}

\usepackage{notations}

\newcommand{\semot}{SemOT}

\usepackage{hyperref}
\usepackage{url}

\usepackage[capitalize,noabbrev]{cleveref}

\usepackage{tikz}
\usetikzlibrary{
    arrows.meta,
    decorations.pathmorphing,
    positioning,
    decorations.shapes,
}

\usepackage[most]{tcolorbox}

\usepackage{multirow}

\usepackage{colortbl}

\usepackage{wrapfig}

\title{Generate in Reconstruction Space, Match in Semantic Space: Transport Geometry for One-Step Generation}

  \author{%
    \textbf{Hugues Van Assel}\thanks{Contact: \texttt{van\_assel.hugues@gene.com}} \quad \textbf{Edward De Brouwer} \quad \textbf{Saeed Saremi} \\[0.2em]
    \textbf{Gabriele Scalia} \quad \textbf{Aviv Regev} \\[0.3em]
    Genentech
  }

\begin{document}

\maketitle

\begin{abstract}
Generative modeling and self-supervised representation learning (SSL) optimize structurally different objectives: generative training rewards distributional fidelity, while SSL rewards semantic coherence. 
Yet recent work repeatedly finds that SSL features improve generative training, though the mechanism of this synergy remains unclear.
Here, we study the benefits of SSL in generative modeling in the framework of one-step generation where the role of representation is explicit: frozen SSL features are used to match generated samples to real data. 
We use the Sinkhorn divergence in that feature space, providing a tractable surrogate for the Wasserstein distance, the population-level discrepancy approximated by Fr\'echet-style evaluation metrics (such as FID). We call this approach \semot{} (Semantic Optimal Transport).
We find that this objective becomes highly effective when computed in a semantically structured SSL feature space (a 39$\times$ reduction in ImageNet FID). 
We trace this behavior primarily to matching estimation: semantic SSL features that suppress nuisance reconstruction details induce a more compact geometry, making distribution matching more tractable.
As a consequence, the best training SSL features need not match the features used by the evaluation metric.
In particular, we show that using Inception as the feature extractor can improve FID while degrading matching stability and sample quality, revealing a form of metric hacking.
Using extensive experiments on ImageNet, we identify which SSL feature families lead to best generation performance and show that matching stability is a quantitative criterion for selecting them.
Code is available at \url{https://github.com/Genentech/semantic-transport-generation}.

\end{abstract}

\section{Introduction}
\label{sec:intro}

Self-supervised representation learning and generative modeling have matured as two central yet historically separate paradigms in AI. Generative modeling aims to learn a data distribution well enough to synthesize new samples: from latent-variable one-step mappings (e.g., VAEs, GANs, normalizing flows) to iterative noise-to-data trajectories in diffusion, score-based, and flow-matching families \citep{kingma2014vae,goodfellow2014gan,dinh2017realnvp,ho2020ddpm,song2021sde,lipman2023flowmatching,liu2023rectifiedflow}. In this paradigm, success is judged by synthesis quality and coverage, because the downstream goal is generation itself.
Self-supervised learning targets a different objective: learning representations that expose semantic structure and invariances useful for downstream perception tasks. Methods range from reconstruction-style objectives (e.g., masked autoencoding) to latent alignment and joint-embedding objectives (e.g., contrastive or self-distillation families) \citep{he2022mae,he2020momentum,grill2020bootstrap,oquab2024dinov2,assel2025joint}. Here, success is measured by downstream transfer rather than sample synthesis: representations are evaluated either frozen (e.g., linear probe or k-NN) or after end-to-end fine-tuning on tasks such as classification, detection, segmentation, depth, and robustness. 

These objectives are formally orthogonal, yet there is growing empirical evidence that aligning generative training with pretrained SSL representations improves performance at multiple levels of the generative pipeline~\citep{yu2024repa,yao2025reconstruction,zheng2025diffusion}.
Depending on the method, the SSL feature extractor serves as a regularizer~\citep{yu2024repa,wang2025haste,wang2025diffuse,singh2025whatmatters}, an alignment target~\citep{yao2025reconstruction,leng2025repae,chen2025aligntok,bi2025vision}, or a generative substrate~\citep{zheng2025diffusion,gao2025one,lei2025advancing}. 
Yet why it helps in each case remains poorly understood.

To study this question we focus on one-step generative models trained by distribution matching, in which a generator maps noise directly to data by minimizing a discrepancy between generated and real samples. Early work formulated this matching in the ambient data space~\citep{arjovsky2017wasserstein,feydy2019interpolating,li2015generative}.
However, these methods have historically been difficult to scale and have generally fallen short of the strongest generative models~\citep{genevay2018learning,li2015generative}.
In this work, we lift the distributional matching objective to a feature space.
Unlike approaches that use the feature extractor as an auxiliary regularizer, here it plays a first-order role: it directly defines the geometry in which generated and real samples are compared, and therefore shapes the training signal. This choice also creates a structural alignment with standard evaluation: Fréchet-style metrics are Gaussian approximations to Wasserstein-type discrepancies in feature space, so training and evaluation measure the same quantity up to the choice of extractor. One-step generation in this framework is therefore an unusually transparent setting for studying what makes a feature extractor effective for generative training.

\begin{figure}[t]
\centering
\begin{tikzpicture}[
    x=0.88cm,
    y=0.76cm,
    font=\normalsize,
    sample/.style={
        draw,
        rounded corners=2pt,
        thick,
        align=center,
        minimum width=2.65cm,
        minimum height=0.62cm,
        fill=white
    },
    adapter/.style={
        draw,
        rounded corners=2pt,
        thick,
        align=center,
        minimum width=1.45cm,
        minimum height=0.50cm,
        draw=black!55,
        fill=black!6
    },
    featureop/.style={
        draw,
        rounded corners=2pt,
        thick,
        align=center,
        minimum width=1.45cm,
        minimum height=0.50cm,
        draw=black!70,
        fill=black!12
    },
    learned/.style={
        draw,
        rounded corners=2pt,
        thick,
        align=center,
        minimum width=1.45cm,
        minimum height=0.50cm,
        draw=orange!55!black,
        fill=orange!16!white
    },
    labelbox/.style={
        align=center,
        font=\small
    },
    arr/.style={
        -{Latex[length=2.6mm,width=1.8mm]},
        line width=1.15pt,
        draw=black!75,
        line cap=round,
        line join=round,
        shorten >=1pt
    },
    ot/.style={
        -{Latex[length=2.5mm,width=1.7mm]},
        line width=1.15pt,
        draw=green!45!black,
        dash pattern=on 4.5pt off 2.6pt,
        line cap=round,
        line join=round,
        shorten >=1pt
    }
]
    \draw[draw=blue!45!black, thick, rounded corners=4pt, fill=blue!6!white]
        (0.0, 5.70) rectangle (15.4, 6.90);
    \draw[draw=orange!55!black, thick, rounded corners=4pt, fill=orange!10!white]
        (0.0, 3.20) rectangle (15.4, 4.40);
    \draw[draw=green!35!black, thick, rounded corners=4pt, fill=green!8!white]
        (0.0, 0.70) rectangle (15.4, 1.90);

    \node[font=\bfseries\normalsize, anchor=south] at (7.7, 6.95) {Ambient space $\gX$};
    \node[font=\bfseries\normalsize, anchor=south] at (7.7, 4.45) {Generation space $\gM$};
    \node[font=\bfseries\normalsize, anchor=south] at (7.7, 1.95) {Feature space $\gF$};

    \node[sample] (x) at (2.4, 6.30) {$\vx \sim p$};
    \node[adapter] (eadapt) at (2.4, 5.05) {$\Enc$};
    \node[sample] (uplus) at (2.4, 3.80) {$\vu^r = \Enc(\vx)$};

    \node[sample] (u0) at (6.7, 3.80) {$\vu_0 \sim p_0$};
    \node[learned] (Ttheta) at (9.8, 3.80) {$\gen_\theta$};
    \node[sample] (u1) at (12.9, 3.80) {$\vu_1 = \gen_\theta(\vu_0)$};
    \node[adapter] (render) at (12.9, 5.05) {$\Dec$};
    \node[sample] (xhat) at (12.9, 6.30) {$\hat{\vx} = \Dec(\vu_1)$};

    \node[featureop] (phiL) at (2.4, 2.55) {$\phi$};
    \node[featureop] (phiR) at (12.9, 2.55) {$\phi$};
    \node[sample] (hplus) at (2.4, 1.30) {$\vh^r = \phi(\vu^r)$};
    \node[sample] (h) at (12.9, 1.30) {$\vh = \phi(\vu_1)$};

    \draw[arr] (x) -- (eadapt);
    \draw[arr] (eadapt) -- (uplus);
    \draw[arr] (u0) -- (Ttheta);
    \draw[arr] (Ttheta) -- (u1);
    \draw[arr] (u1) -- (render);
    \draw[arr] (render) -- (xhat);
    \draw[arr] (uplus) -- (phiL);
    \draw[arr] (phiL) -- (hplus);
    \draw[arr] (u1) -- (phiR);
    \draw[arr] (phiR) -- (h);

    \draw[ot] (h.west) --
        node[midway, above=-3pt, labelbox] {$\hat T_\varepsilon^{\mathrm{cfg}}(\vh)$ (Sinkhorn CFG)}
        (hplus.east);
\end{tikzpicture}
\caption{Overview of \semot{}. A prior state $\vu_0 \sim p_0$ is mapped in one step by $\gen_\theta$ to a generated state $\vu_1$ in generation space $\gM$, then decoded to the ambient sample $\hat{\vx}=\Dec(\vu_1)$. In parallel, a real datum $\vx \sim p$ is encoded into the generation-space target $\vu^r=\Enc(\vx)$. Both states are embedded by the frozen SSL feature extractor $\phi$ into feature vectors $\vh=\phi(\vu_1)$ and $\vh^r=\phi(\vu^r)$, and the training loss is a Sinkhorn transport problem in feature space $\gF$. Thus, generation is performed in generation space, while the transport geometry used for training is defined in feature space. In practice, the loss is evaluated over minibatches; the figure shows the same construction schematically for a single pair of states. In input-space models, $\gM=\gX$ and $\Enc=\Dec=\mathrm{Id}$. In latent-space models, applying $\phi$ directly on the compact generation space $\gM$ (e.g., a VAE latent) avoids decoding to pixels; in our experiments, $\phi$ is obtained by training from scratch or distilling ambient feature extractors (DINOv3, Inception) into generation space.}
\label{fig:feature_geometry_setup}
\end{figure}
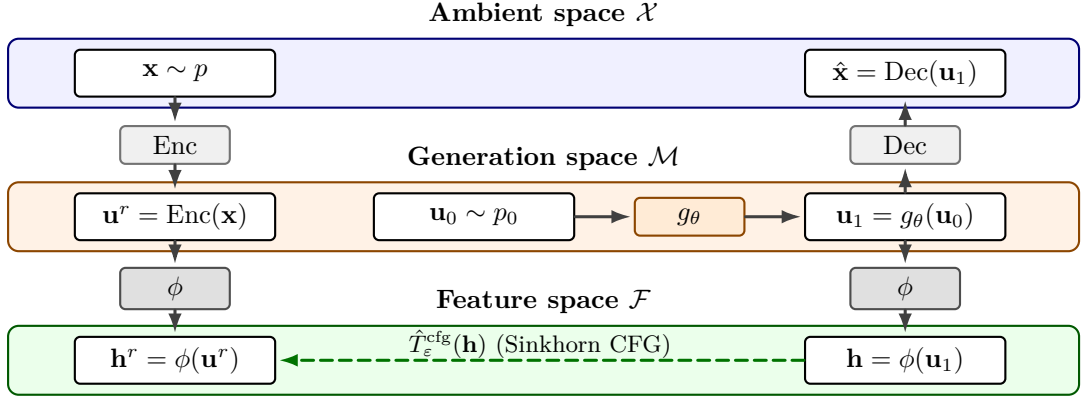

\paragraph{Contributions.}
Our approach is motivated by the intuition that \textbf{semantically rich feature spaces are geometrically better suited to distribution matching than reconstruction-oriented spaces}.
By discarding nuisance reconstruction variation and concentrating on semantic variation, these features yield a lower-dimensional problem with better statistical properties.
At a high level, our approach breaks down the challenging problem of learning to sample from a high-dimensional data distribution into two stages (\Cref{fig:feature_geometry_setup}): we generate samples in an auto-encoding reconstruction space, but set up the distributional matching objective in a semantic space built on top of it, which by design has a better geometry (\Cref{fig:transport_plan}). This separation of generation and matching into two spaces with distinct geometric roles is the core conceptual idea of this work. Our contributions are as follows:

\begin{itemize}[leftmargin=2em, itemsep=3pt, topsep=2pt]
    \item \textbf{Semantic features unlock competitive one-step generation.} We perform matching in a frozen SSL feature space using the Sinkhorn divergence, a discrepancy that interpolates between Wasserstein distance and MMD, while generating in the reconstruction-oriented latent space of a pretrained VAE (\Cref{fig:feature_geometry_setup}); we call this framework \semot{} (Semantic Optimal Transport).
    One-step Sinkhorn-based generation without a semantic feature extractor is severely limited~\citep{genevay2018learning}; we show that SSL features built directly in the reconstruction latent space reduce FID-Inception by $39\times$ on class-conditional ImageNet (\Cref{tab:imagenet_results}), closing much of the gap to multi-step diffusion and flow-matching models for a generator trained from scratch, via a classifier-free guidance formulation for Sinkhorn training (\Cref{sec:sinkhorn_method}). We provide precise recipes for obtaining such features, either by training SSL models directly on VAE latents (\Cref{app:mae_pretrain}) or by distilling pretrained SSL models such as DINOv3 into the VAE latent space (\Cref{app:dinov3_distill}), and compare both approaches in \Cref{sec:exp_imagenet}.
    \item \textbf{Matching-estimation stability as a key criterion.} The dramatic improvement raises the question of what makes an SSL feature extractor effective in this framework. We identify transport estimation quality as the primary driver: features with lower intrinsic dimensionality induce more stable OT transport plans within the Sinkhorn divergence and more reliable training signals (\Cref{fig:transport_plan}). We derive the OT stability metric $\tilde{D}_N$ (\Cref{sec:eval_metrics}), which directly predicts generation quality across extractor families (\Cref{tab:imagenet_results}) and provides a computation-light selection criterion without full training runs. As a direct consequence, the optimal training feature extractor does not always match the evaluation feature extractor: geometric alignment with the evaluation metric and OT estimation stability are two distinct criteria that can point to different extractors, and optimizing for the former alone can hurt training (\Cref{sec:feature_geometry}).
    \item \textbf{Evaluation geometry exposes metric limitations.} When the training feature extractor coincides with the evaluation feature extractor, Sinkhorn training optimizes a discrepancy structurally aligned with the evaluation metric, enabling a form of metric hacking. In particular, using Inception as the training feature extractor directly targets the same feature space as FID, yielding competitive Fr\'{e}chet scores while producing perceptually inferior samples (\Cref{fig:encoder_comparison}, \Cref{tab:imagenet_results}). Beyond our specific framework, this highlights a broader weakness of widely used feature-based metrics in generative modeling: strong performance on FID does not imply better sample quality.

\end{itemize}

\section{Background}
\label{sec:background}

This section introduces the background material for this work. We start with optimal transport and the Sinkhorn divergence, which formalize distribution matching in feature space and provide our practical training objective. We also cover prior work on statistical estimation of OT maps from finite samples~\citep{pooladian2024entropicestimationoptimaltransport}. We close with an overview of prior works leveraging SSL representations in generative models.

\paragraph{Entropic OT and Sinkhorn divergence.}\label{sec:sinkhorn_div}
We leverage optimal transport to compare generated and target distributions in the feature space where the cost is defined. For probability measures $P,Q \in \mathcal{P}_2(\gF)$ on a feature space $\gF$, the quadratic transport objective is
\begin{equation}
\label{eq:w2_background}
W_2^2(P,Q)
\defeq
\min_{\pi\in\Pi(P,Q)}
\int_{\gF\times\gF}
\|\vh-\vh'\|^2\,d\pi(\vh,\vh').
\end{equation}
Here $\Pi(P,Q)$ denotes the set of couplings between $P$ and $Q$. $W_2$ is a proper metric on the space of probability measures: unlike likelihood-based divergences, it remains finite and provides a useful training signal even when generated and real distributions have disjoint support, making it a natural population-level objective for generative modeling~\citep{villani2008,arjovsky2017wasserstein,genevay2018learning}. Given generic empirical measures $\alpha = \frac{1}{N}\sum_{i=1}^N \delta_{\vh_i}$ and $\beta = \frac{1}{N}\sum_{j=1}^N \delta_{\vh'_j}$ on $\gF$, entropic regularization provides a tractable finite-sample approximation to the quadratic Wasserstein problem: it smooths the coupling problem with an entropy penalty, is computable by Sinkhorn iterations, and is well suited to parallel GPU implementation~\citep{cuturi2013sinkhorn,peyre2019computational}:
\begin{equation}
\label{eq:eot}
    \mathrm{OT}_\varepsilon(\alpha, \beta) = \min_{\vpi \in \Pi(\alpha, \beta)} \sum_{i,j} \pi_{ij} \, C_{ij} + \varepsilon \sum_{i,j} \pi_{ij} \log \pi_{ij}\,,
\end{equation}
where $C_{ij} = \|\vh_i - \vh'_j\|^2$ and $\Pi(\alpha,\beta)$ is the set of coupling matrices with marginals $\alpha$ and $\beta$. However, $\mathrm{OT}_\varepsilon(\alpha,\alpha) > 0$ even when both arguments coincide, causing entropic bias and shrinkage when used as a training loss. 
The \emph{Sinkhorn divergence}~\citep{genevay2018learning,feydy2019interpolating} removes this bias:
\begin{equation}
S_\varepsilon(\alpha, \beta)
=
\mathrm{OT}_\varepsilon(\alpha, \beta)
- \tfrac{1}{2}\,\mathrm{OT}_\varepsilon(\alpha, \alpha)
- \tfrac{1}{2}\,\mathrm{OT}_\varepsilon(\beta, \beta).
\end{equation}
It satisfies $S_\varepsilon(\alpha,\alpha)=0$, metrizes weak convergence, and interpolates between unregularized OT and MMD~\citep{feydy2019interpolating}. In our setting, Sinkhorn is therefore the practical surrogate to the ideal quadratic Wasserstein objective. 
\paragraph{Matching estimation.}
In practice, the Sinkhorn divergence is computed over finite mini-batches rather than the full distributions. From each mini-batch, the Sinkhorn solver produces a coupling $\vpi$ (see \Cref{app:sinkhorn_details}) that assigns to each generated feature $\vh_i$ a target indicating where it should move to better match the real distribution; this per-point target is the matching signal that drives generator training. A key question is then how faithfully this finite-sample signal tracks the ideal population-level matching map: poor estimation quality translates directly into noisy and unreliable training gradients, regardless of the quality of the Sinkhorn objective itself.
The \emph{entropic barycentric estimator} maps each source point to its barycenter under the optimal coupling $\vpi$: $\hat{T}_\varepsilon(\vh_i) = N\sum_{j=1}^N \pi_{ij}\,\vh'_j$.
Under regularity assumptions on the feature marginals, this estimator converges to the ideal population Monge map $T_0$, the unique map $T_0\colon\gF\to\gF$ pushing $P$ forward to $Q$ minimizing $\mathbb{E}_P[\|T_0(\vh)-\vh\|^2]$ (Brenier's theorem)~\citep{pooladian2024entropicestimationoptimaltransport}, with mean-squared estimation error (in $L^2$ under the source population measure) decaying as
\begin{equation}
\label{eq:pnw_rate}
\E \bigl\|\hat T_\varepsilon - T_0\bigr\|_{L^2}^2 \;\lesssim\; N^{-\tfrac{s+2}{2(d+s+2)}} \log N\,,
\end{equation}
where $d$ is the intrinsic dimensionality of the feature distribution and $s \leq 2$ the Hölder smoothness of the source and target feature densities. The rate is controlled primarily by $d$: more compact feature distributions yield faster estimation and more reliable matching signals from finite minibatches. This is the statistical bridge we use later: selecting SSL feature extractors that reduce $d$ directly improves matching estimation quality.

\paragraph{SSL for generation.}
There is growing empirical evidence that aligning generative training with pretrained SSL representations improves performance across multiple levels of the generative pipeline.
At the generator level, REPA~\citep{yu2024repa} shows clear optimization gains by regularizing the denoiser's intermediate representations to align with clean features from a frozen pretrained visual featurizer; follow-up work refines the alignment schedule~\citep{wang2025haste}, proposes dispersive representation regularization~\citep{wang2025diffuse}, and probes which SSL featurizer properties drive alignment gains~\citep{singh2025whatmatters}.
At the encoder level, recent work aligns or redesigns latent spaces using foundation representations, including end-to-end tuning and alignment of autoencoders~\citep{yao2025reconstruction,leng2025repae,chen2025aligntok,bi2025vision}.
Beyond regularization and encoder alignment, some approaches use SSL representations as a generative substrate, generating directly in an SSL feature space~\citep{zheng2025diffusion,gao2025one,lei2025advancing}; these require learning a decoder from that space back to the pixel domain, which is a non-trivial additional task. Among prior work, the drifting approach~\citep{deng2026generative} is most closely related to ours: it also performs one-step generation and leverages SSL features, but approximates the transport cost via a one-round double-softmax rather than solving a proper matching problem. Grounding the matching cost in the Sinkhorn divergence allows direct application of the matching estimation theory above~\citep{pooladian2024entropicestimationoptimaltransport}, enabling a more transparent study of the SSL featurizer's effect on generation quality; see \Cref{app:drifting_details} for a detailed comparison with drifting. Distributional losses also help beyond one-step models: with a kernel scoring rule, an MMD-type objective, in place of the regression loss, diffusion models learn the full posterior of clean data and can sample in few steps~\citep{debortoli2025distributional}. Concurrently, FD-loss~\citep{yang2026representation} post-trains generators with a Fr\'echet distance in representation space, i.e., the $W_2$ distance between Gaussian approximations of the feature distributions. \semot{} instead trains a one-step generator from scratch with the Sinkhorn divergence, which compares full feature distributions at the cost of estimating them from minibatches, the matching estimation problem we study.
Across all these approaches, why SSL features help generation remains poorly understood: the gains are empirical, and it is unclear which properties of the representations drive the performance gains.

\section{Feature Geometry for One-Step Generation}
\label{sec:method}

This section formalizes \semot{}, one-step generation in a feature space defined by a frozen SSL extractor (\Cref{fig:feature_geometry_setup}). \Cref{sec:feature_geometry} introduces the generation setup and decomposes the gap between the evaluation objective and the training loss, identifying two featurizer-dependent quantities: geometric alignment with the evaluation featurizer, and matching estimation quality from finite minibatches. \Cref{sec:sinkhorn_method} then derives the Sinkhorn training objective with classifier-free guidance.

\subsection{Feature Geometry and the Evaluation-Training Gap}
\label{sec:feature_geometry}

\paragraph{Generation space.} Let $\gX$ denote the ambient data space and $p$ the data distribution on $\gX$. Generation is performed in a \emph{generation space} $\gM$ equipped with an encoder $\Enc:\gX\to\gM$ and a decoder $\Dec:\gM\to\gX$. Following latent diffusion~\citep{rombach2022ldm}, $\gM$ is typically a compact latent space (e.g.\ a VAE latent), in which generation is more tractable than in $\gX$ directly.
A one-step generator $\gen_\theta:\gM\to\gM$ maps noise $\vu_0\sim p_0$ (typically standard Gaussian) to a generated point $\gen_\theta(\vu_0)$, which the decoder maps back to $\gX$. The generator is trained by matching generated points to encoded real data $\vu^r \defeq \Enc(\vx)\in\gM$, $\vx \sim p$.

\paragraph{Matching cost in feature space.} The defining choice of \semot{} is to measure the discrepancy between generated and real samples not in $\gM$ or $\gX$, but in a separate \emph{feature space} $\gF$ with different geometric properties, designed to be more suitable for matching estimation. Concretely, a frozen SSL feature extractor $\phi \colon \gM \to \gF$ is applied directly to points in generation space, avoiding a round-trip through the decoder to ambient space and keeping feature extraction computationally efficient. The transport cost is the squared feature distance $\|\phi(\vu)-\phi(\vu')\|^2$.
The featurizer $\phi$ therefore determines both the population matching geometry and the discrete Sinkhorn problems used to approximate it in practice. We write $\phi_\star \colon \gM \to \gF_\star$ for the evaluation featurizer, which defines the geometry in which generation quality is measured.

\paragraph{Wasserstein objectives.}
For a generator $g:\gM\to\gM$, we define the generated and real feature probability measures
\begin{equation}
q_{\phi,g}
\defeq
(\phi\circ g)_\# p_0,
\qquad
r_\phi
\defeq
(\phi\circ\Enc)_\# p,
\end{equation}
and their corresponding evaluation-space feature probability measures
\begin{equation}
q_{\star,g}
\defeq
(\phi_\star\circ g)_\# p_0,
\qquad
r_\star
\defeq
(\phi_\star\circ\Enc)_\# p.
\end{equation}
Matching generated and real measures in each geometry gives the Wasserstein objectives (see \Cref{sec:background})
\begin{equation}
J_\phi(g)
\defeq
W_2^2(q_{\phi,g},r_\phi),
\qquad
J_\star(g)
\defeq
W_2^2(q_{\star,g},r_\star).
\end{equation}
$J_\phi$ is the population-level training objective, optimized in the geometry of $\phi$; $J_\star$ is the evaluation objective, measuring discrepancy in the geometry of $\phi_\star$. The central question is how well minimizing $J_\phi$ via its empirical Sinkhorn approximation drives $J_\star$ toward zero.

\paragraph{Empirical Sinkhorn objective.}
$J_\phi$ is not directly accessible, so training uses an empirical Sinkhorn approximation. Given minibatch samples $\vu_{0,1},\ldots,\vu_{0,N}\sim p_0$ and $\vx_1,\ldots,\vx_N\sim p$, define the empirical feature measures $\hat q_{\phi,g} \defeq \frac{1}{N}\sum_{i=1}^N \delta_{\phi(g(\vu_{0,i}))}$ and $\hat r_\phi \defeq \frac{1}{N}\sum_{j=1}^N \delta_{\phi(\Enc(\vx_j))}$, and the one-batch objective optimized in training:
\begin{equation}
\hat{J}_{\phi,\varepsilon,N}(g)
\defeq
S_\varepsilon(\hat q_{\phi,g},\hat r_\phi).
\end{equation}
Measured against this surrogate, the evaluation objective splits exactly into the training loss and two featurizer-dependent gaps:
\begin{tcolorbox}[colframe=white!, top=2pt, left=2pt, right=2pt, bottom=2pt]
\begin{center}
\textbf{Evaluation-Training Decomposition}
\begin{equation}
\label{eq:theory_main_bound}
J_{\star}(g)
\;=\;
\underbrace{\hat{J}_{\phi,\varepsilon,N}(g)}_{\text{training loss}}
\;+\;
{\color{colorGeom}\underbrace{J_{\star}(g)-J_{\phi}(g)}_{\Delta^\mathrm{geom}_\phi(g)\text{: geometric misalignment}}}
\;+\;
{\color{colorEst}\underbrace{J_{\phi}(g)-\hat{J}_{\phi,\varepsilon,N}(g)}_{\Delta^\mathrm{est}_{\phi,\varepsilon,N}(g)\text{: matching estimation}}}.
\end{equation}
\end{center}
\end{tcolorbox}
The training loss $\hat{J}_{\phi,\varepsilon,N}(\gen_\theta)$ also depends on $\phi$: the generator is trained through the composition $\phi\circ\gen_\theta$, and a smoother feature map makes this target easier to approximate with a finite-capacity network and yields better-behaved gradients through the Jacobian of $\phi$. We do not quantify this optimization effect here.
The two gaps, in contrast, each depend on a distinct property of the featurizer and can both be measured without training a generator; we detail them next.

\paragraph{\geomcolor{Geometric misalignment.}}
The term $\textcolor{colorGeom}{\Delta^\mathrm{geom}}$ measures whether the feature geometry of $\phi$ preserves the matching problem seen by the evaluation featurizer $\phi_\star$. When it is small, improving the training objective also improves the evaluation objective, so this axis captures semantic faithfulness. In the experiments, we measure it by how well $\phi_\star$ can be predicted linearly from $\phi$ (\Cref{fig:encoder_alignment}). In the ideal case where $\phi_\star = A\circ\phi$ on the relevant supports for some isometric embedding $A:\gF\to\gF_\star$, the two featurizers induce the same matching geometry, so $J_\star(g)=J_\phi(g)$ for every generator $g$ and $\Delta^\mathrm{geom}_\phi=0$.

\paragraph{\estcolor{Matching estimation.}}
The term $\textcolor{colorEst}{\Delta^\mathrm{est}}$ measures whether the entropic minibatch training signal under $\phi$ tracks the population quadratic Wasserstein problem. Training sees this gap only through the gradient of the empirical Sinkhorn loss with respect to the generated features. For a fixed generator $\gen_\theta$ and a minibatch $\{(\vu_{0,i},\vx_i)\}_{i=1}^N$, let $\vh_i=\phi(\gen_\theta(\vu_{0,i}))$ and $\vh_j^r=\phi(\Enc(\vx_j))$. Then
\begin{equation}
\label{eq:theory_gradient_bridge}
\nabla_{\vh_i}\hat{J}_{\phi,\varepsilon,N}(\gen_\theta)
=
-\frac{2}{N}\Bigl(\hat{T}_\varepsilon^{q,r}(\vh_i)-\hat{T}_\varepsilon^{q,q}(\vh_i)\Bigr),
\end{equation}
where $\vpi^{q,r}$ and $\vpi^{q,q}$ are the optimal couplings of $\mathrm{OT}_\varepsilon(\hat{q}_{\phi,\gen_\theta},\hat r_\phi)$ and $\mathrm{OT}_\varepsilon(\hat{q}_{\phi,\gen_\theta},\hat{q}_{\phi,\gen_\theta})$ respectively, and $\hat{T}_\varepsilon^{q,r}(\vh_i) \defeq N\sum_j \pi^{q,r}_{ij}\vh_j^r$, $\hat{T}_\varepsilon^{q,q}(\vh_i) \defeq N\sum_k \pi^{q,q}_{ik}\vh_k$ are the corresponding cross- and self-barycentric projections (see \Cref{app:sinkhorn_details} for the log-domain algorithms used to compute these). 
The update direction is thus exactly the cross barycentric map minus the self-transport correction. Under regularity assumptions on the feature marginals, the empirical cross barycentric map converges to the population Monge map~\citep{pooladian2024entropicestimationoptimaltransport} at the rate given in~\eqref{eq:pnw_rate}, governed by the intrinsic dimensionality $d$ of the features (proxied in practice by the effective rank of the feature covariance; \Cref{sec:eval_metrics}).
This makes the stability of these targets across minibatches a natural proxy for matching estimation: if they vary strongly between target subsamples, the Sinkhorn updates do not yet track a stable approximation of the population transport problem. We measure this stability with $\tilde{D}_N$, defined in~\eqref{eq:ot_stability}.

\paragraph{Implications for featurizer choice.}
Two measurable properties of a featurizer matter: its alignment with $\phi_\star$ (\Cref{fig:encoder_alignment}) and the stability of its minibatch transport targets ($\tilde{D}_N$), which need not agree.
We expect, and test in \Cref{sec:experiments}, that
(i)~a semantic featurizer outperforms matching directly in $\gM$, as it can reduce both gaps;
(ii)~a featurizer aligned with $\phi_\star$ but with unstable targets improves the metric of $\phi_\star$ without improving samples, a form of metric hacking that independent evaluation encoders reveal;
and (iii)~when $\textcolor{colorGeom}{\Delta^\mathrm{geom}}=0$ by construction, featurizers rank by matching stability alone.

\subsection{Sinkhorn Training with Classifier-Free Guidance}
\label{sec:sinkhorn_method}

\semot{} trains $\gen_\theta$ for class-conditional generation by minimizing the Sinkhorn divergence between generated and real feature distributions in SSL feature space. Write $q_{\phi,\theta}(\cdot|c,w) \defeq (\phi\circ\gen_\theta(\cdot|c,w))_\# p_0$ for the generated feature distribution at class label $c$ and guidance weight $w$, $r_\phi^c \defeq (\phi\circ\Enc)_\# p(\cdot|c)$ for the class-conditional real feature distribution, and $r_\phi^u \defeq (\phi\circ\Enc)_\# p$ for the class-marginal (unconditional) real feature distribution; their minibatch empirical counterparts are $\hat{q}_{\phi,\theta}(\cdot|c,w)$, $\hat{r}_\phi^c$, $\hat{r}_\phi^u$. In contrast to standard diffusion CFG, which is applied at inference by mixing conditional and unconditional predictions at the same latent state, we incorporate guidance directly into training by keeping the generated distribution fixed and interpolating between Sinkhorn objectives toward the conditional and unconditional real distributions:
\begin{tcolorbox}[colframe=white!, top=2pt, left=2pt, right=2pt, bottom=2pt]
\begin{center}
\textbf{\semot{} Training Objective (Sinkhorn CFG)}
\begin{equation}
\label{eq:cfg_objective}
    \min_{\theta} \quad
    (1+w)\, S_\varepsilon\!\big(\hat{q}_{\phi,\theta}(\cdot|c,w),\, \hat{r}_\phi^c\big)
    \;-\; w\, S_\varepsilon\!\big(\hat{q}_{\phi,\theta}(\cdot|c,w),\, \hat{r}_\phi^u\big)
\end{equation}
\end{center}
\end{tcolorbox}
where $w \ge 0$ is the guidance weight: $w = 0$ recovers the conditional Sinkhorn divergence, while $w > 0$ adds a repulsion from the class-marginal data, so that the objective is no longer a divergence.
By the envelope theorem, the gradient can be computed with the Sinkhorn couplings held at their optimal values, so no backpropagation through the solver is required; with respect to the generated features, it equals $-\frac{2}{N}\,\hat{T}_\varepsilon^{\mathrm{cfg}}(\vh_i)$, where
\begin{equation}
\label{eq:cfg_field}
    \hat{T}_\varepsilon^{\mathrm{cfg}}(\vh_i)
    \;=\; (1+w)\,\hat{T}_\varepsilon^{q,c}(\vh_i)
    \;-\; w\,\hat{T}_\varepsilon^{q,u}(\vh_i)
    \;-\; \hat{T}_\varepsilon^{q,q}(\vh_i)\,,
\end{equation}
and $\hat{T}_\varepsilon^{q,c}(\vh_i)$, $\hat{T}_\varepsilon^{q,u}(\vh_i)$ are the conditional and class-marginal cross-transport barycenters. Grouping the cross terms as $\hat{T}_\varepsilon^{q,c} + w\,(\hat{T}_\varepsilon^{q,c} - \hat{T}_\varepsilon^{q,u})$ mirrors the standard CFG decomposition: guidance extrapolates from the class-agnostic toward the class-conditional transport target. Since both targets are barycenters of real samples, the update stays anchored to the data despite the signed objective; as in standard CFG, too large a $w$ overshoots, so $w$ is selected per model (\Cref{sec:exp_imagenet}). Three Sinkhorn solves are needed per step: conditional cross, class-marginal cross, and self-transport; since both divergences share the same generated distribution, their self-transport terms coincide and contribute a single $\hat{T}_\varepsilon^{q,q}$ term.
At inference, choose a class label $c$ and guidance weight $w$, sample $\vu_0 \sim p_0$, map $\vu_1 = \gen_\theta(\vu_0 \mid c,w)$, and decode $\hat\vx = \Dec(\vu_1)$.

\section{Experiments}
\label{sec:experiments}

\begin{figure*}[t]
\centering
\includegraphics[width=\linewidth]{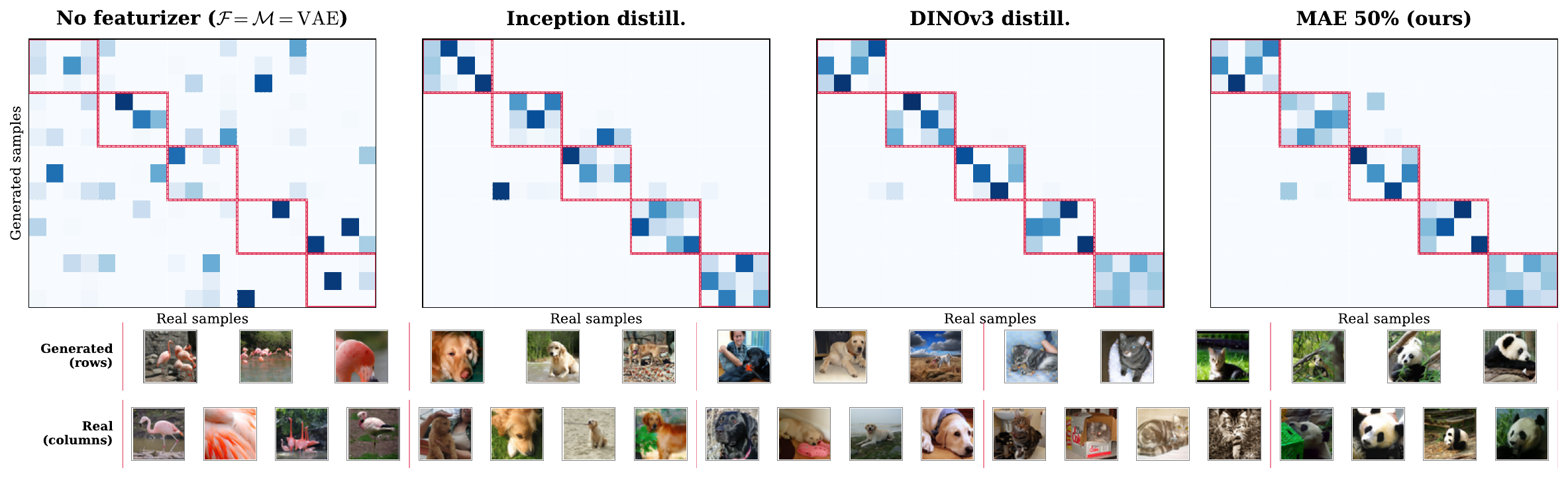}
\caption{%
  Sinkhorn coupling matrices $\pi^\varepsilon$ between generated samples (rows) and real samples (columns) for five ImageNet classes, including a confusable pair (golden retriever, Labrador).
  Generated samples are fixed across all panels (from the MAE $50\%$ model); only the featurizer $\phi$ varies, defining the transport cost $C_{ij} = \|\phi(\vu_i) - \phi(\vu_j^+)\|^2$.
  Red outlines indicate the expected same-class blocks along the diagonal.
  Without a semantic featurizer (left), VAE features alone fail to recover the block-diagonal structure.
  Inception distillation partially recovers the block-diagonal structure, while DINOv3 distillation and MAE $50\%$ produce sharper same-class matching.
  Coupling computed with $\varepsilon = 0.05 \cdot \mathrm{std}(C)$ and 50 Sinkhorn iterations.%
}
\label{fig:transport_plan}
\end{figure*}

We train \semot{} for class-conditional generation, varying the frozen SSL feature extractor~$\phi$ and measuring generation quality and featurizer diagnostics with the metrics below. The main experiments cover ImageNet image generation; experiments with single-cell RNA-Seq can be found in \Cref{app:scrna}.

\subsection{Evaluation metrics}
\label{sec:eval_metrics}

\paragraph{Fr\'echet metrics.}
The standard quality metric in generative modeling, across both natural images~\citep{heusel2017gans} and single-cell genomics, is the Fr\'echet distance in the feature space~$\gF_\star$ of a fixed evaluation featurizer~$\phi_\star$. Writing $\boldsymbol{\mu}_r, \Sigmab_r$ and $\boldsymbol{\mu}_f, \Sigmab_f$ for the mean and covariance of $r_\star$ and $q_{\star,\gen_\theta}$ respectively,
\begin{equation}
\label{eq:frechet}
\mathrm{FD}_{\phi_\star}
\;=\;
\|\boldsymbol{\mu}_r - \boldsymbol{\mu}_f\|^2
+ \mathrm{tr}\!\left(\Sigmab_r + \Sigmab_f - 2\left(\Sigmab_r^{1/2}\Sigmab_f\Sigmab_r^{1/2}\right)^{1/2}\right),
\end{equation}
which equals $W_2^2(\mathcal{N}(\boldsymbol{\mu}_r,\Sigmab_r), \mathcal{N}(\boldsymbol{\mu}_f,\Sigmab_f))$~\citep{dowson1982frechet}, a lower bound on $W_2^2(r_\star,\,q_{\star,\gen_\theta})=J_\star(\gen_\theta)$~\citep{gelbrich1990formula}; hence $\mathrm{FD}_{\phi_\star} \le J_\star(\gen_\theta)$, a computable lower bound on the evaluation objective in~\eqref{eq:theory_main_bound}. We report $\mathrm{FD}_{\phi_\star}$ for multiple evaluation featurizers, probing featurizer--metric alignment across feature spaces.
Since a Fr\'echet distance folds fidelity and diversity into one number, \Cref{app:independent_eval} complements it with KID~\citep{binkowski2018demystifying}, precision and recall~\citep{kynkaanniemi2019improved}, and density and coverage~\citep{naeem2020reliable}.

\paragraph{OT matching estimation stability.}
We measure the normalized barycentric disagreement
\begin{equation}
\label{eq:ot_stability}
\tilde{D}_N
\;\defeq\;
\frac{1}{C_\phi}
\cdot
\frac{1}{N}\sum_{i=1}^{N}
\bigl\|\hat T_{\varepsilon,\mathcal{A}}^{q,r}(\vh_i) - \hat T_{\varepsilon,\mathcal{B}}^{q,r}(\vh_i)\bigr\|^2,
\end{equation}
where $\mathcal{A}$, $\mathcal{B}$ are two independent size-$N$ draws from a held-out target pool (mimicking two different training minibatches), $\hat T_{\varepsilon,\mathcal{A}}^{q,r}(\vh_i) = N\sum_{j\in\mathcal{A}}\pi^\varepsilon_{ij}\vh_j^r$ is the resulting Sinkhorn cross-transport barycentric map, and $C_\phi$ is the empirical variance of target features, making $\tilde{D}_N$ scale-invariant across featurizer families.
A small $\tilde{D}_N$ means empirical transport targets are stable across minibatch resamples; a large value indicates erratic training directions and a larger matching estimation gap.

\paragraph{Effective rank.}
The effective rank $\exp\!\bigl(-\sum_i \bar\lambda_i \log \bar\lambda_i\bigr)$, where $\bar\lambda_i \defeq \lambda_i/\sum_j\lambda_j$ and $\{\lambda_i\}$ are the eigenvalues of the target feature covariance, is the Vendi score~\citep{friedman2022vendi} of the features under a linear kernel on centered features, a spectral diversity measure used to evaluate generative models; here it proxies the intrinsic dimensionality of the feature distribution; by the estimation rate in \eqref{eq:pnw_rate}~\citep{pooladian2024entropicestimationoptimaltransport}, lower intrinsic dimensionality implies faster OT map estimation.

\subsection{Setup}
\label{sec:exp_imagenet}
We study class-conditional generation on ImageNet at $256 \times 256$ resolution.
Following latent diffusion~\citep{rombach2022ldm}, generation is performed in the latent generation space $\gM \cong \R^{4 \times 32 \times 32}$ of the frozen Stable Diffusion VAE~\citep{rombach2022ldm}:\footnote{\texttt{stabilityai/sd-vae-ft-mse} on Hugging Face.} the encoder $\Enc: \gX \to \gM$ maps real images to latent targets $\vu^r = \Enc(\vx)$ for $\vx \sim p$, and the decoder $\Dec: \gM \to \gX$ maps generated latents back to pixels.
The generator $\gen_\theta$ is DriftDiT-B/2~\citep{peebles2023scalable}: a DiT-Base transformer with patch size~2 operating on $32 \times 32$ latents.
We train with AdamW ($\mathrm{lr} = 4 \times 10^{-4}$), using $(N_\text{pos}, N_\text{neg}, N_\text{unc}) = (128, 64, 32)$ positive, generated/self, and unconditional samples per class, and a cosine schedule over 400k steps on 8 GPUs.
We evaluate with four instances of $\mathrm{FD}_{\phi_\star}$ (Eq.~\ref{eq:frechet}), each corresponding to a different evaluation featurizer~$\phi_\star$ applied to 50k generated samples: \textbf{FID-Inception}~\citep{heusel2017gans} (InceptionV3 pool3, 2048-dim); \textbf{FD-DINOv3}~\citep{simeoni2025dinov3} (CLS token of a large DINO-family ViT, 4096-dim); \textbf{FD-VAE}~\citep{rombach2022ldm} (SD-VAE latent scaled by $0.18215$, 4096-dim), which measures distributional match in the generator's native generation space~$\gM$; and \textbf{FD-CLIP}~\citep{radford2021clip} (CLIP ViT-L/14 image embedding, 768-dim), an evaluation featurizer involved neither in training nor in distillation. \Cref{app:independent_eval} adds FD-DINOv2~\citep{oquab2024dinov2}, the backbone recommended for FD evaluation~\citep{stein2023exposing}, which we leave out of the main table because DINOv2 belongs to the same model family as the DINOv3 teacher.
Offline post-hoc evaluation selects the CFG weight~$w$ per model and reports all metrics at that operating point.

\paragraph{Latent SSL feature extractors.}
\begin{wrapfigure}{r}{0.41\linewidth}
\vspace{-6pt}
\centering
\includegraphics[width=\linewidth]{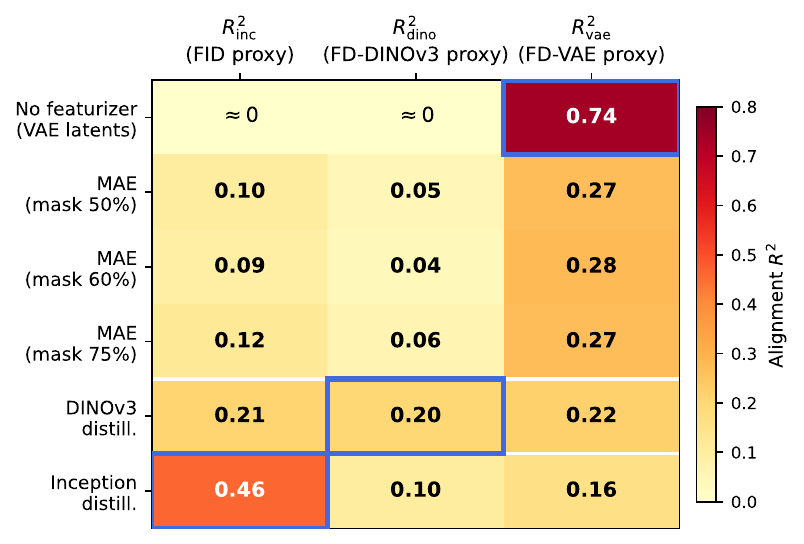}
\caption{%
  Featurizer--evaluation alignment $R^2$ for the six featurizers in \Cref{tab:imagenet_results}.
  Each cell is the held-out $R^2$ of a linear map $\phi(x)\to\phi_\star(x)$ fitted on
  2048 ImageNet-256 latents (methodology in \Cref{app:encoder_alignment}).
  Higher values indicate smaller geometric misalignment ($\Delta^\mathrm{geom}$).%
}
\label{fig:encoder_alignment}
\vspace{-7pt}
\end{wrapfigure}
The matching geometry is determined by a frozen SSL feature extractor $\phi \colon \gM \to \gF$
applied directly to SD-VAE latents; all featurizers use a ViT backbone (32~layers, hidden dimension~$1280$, 16~heads, patch size~$2$) with attention-key facet feature extraction (\Cref{app:key_facet}).
We evaluate three families:
\begin{itemize}[leftmargin=1.5em, itemsep=0pt, topsep=2pt]
  \item \textbf{MAE (mask $50\%$ / $60\%$ / $75\%$).}
        ViT pretrained with masked autoencoding~\citep{he2022mae} directly on SD-VAE latents
        via token-level MSE reconstruction at three mask ratios
        (details in \Cref{app:mae_pretrain}).
  \item \textbf{DINOv3 distillation.}
        ViT student trained on SD-VAE latents to match the CLS and patch tokens of a frozen
        DINOv3 ViT-7B teacher~\citep{simeoni2025dinov3} via cosine similarity loss
        (details in \Cref{app:dinov3_distill}).
  \item \textbf{Inception distillation.}
        ViT student trained on SD-VAE latents to match the features of a frozen
        InceptionV3~\citep{heusel2017gans} (the same model used for FID) via cosine
        similarity loss (details in \Cref{app:inception_distill}).
\end{itemize}

\subsection{Results and Analysis}
\Cref{tab:imagenet_results} reports generation quality and featurizer diagnostics for all families alongside the no-featurizer baseline.

\begin{table*}[t]
\centering
\caption{%
  Unified featurizer comparison on ImageNet class-conditional generation (default backbone: 32-layer ViT).
  All rows are \semot{} (\Cref{sec:sinkhorn_method}) trained with the listed featurizer~$\phi$.
  Metrics as defined in \Cref{sec:eval_metrics}; FD-CLIP uses an evaluation featurizer involved neither in training nor in distillation (further metrics in \Cref{app:independent_eval}).
  \textbf{Bold}: best per column; \underline{underline}: second best.
}
\label{tab:imagenet_results}
\setlength{\tabcolsep}{3pt}
\begin{tabular*}{\textwidth}{@{\extracolsep{\fill}}lcccccc@{}}
\toprule
Featurizer & FID-Inc$\downarrow$ & FD-DINOv3$\downarrow$ & FD-VAE$\downarrow$ & FD-CLIP$\downarrow$ & Eff.\ rank & $\tilde{D}_{128}{\downarrow}$ \\
\midrule
No featurizer ($\gM=\gF$) & 134.56            & 90.45                 & 179.80                & 195.85 & 79.0 & 0.194 \\
\midrule
Inception distill. & 4.88              & 68.84                 & 510.26                & 78.19 & 87.0 & 0.240 \\
\midrule
DINOv3 distill.    & \underline{3.60}  & \underline{43.87}     & \underline{129.31}    & \textbf{17.07} & 68.0 & 0.191 \\
\midrule
MAE (mask $75\%$)  & 6.66              & \textbf{40.87}        & 161.38                & \underline{25.87} & 57.3 & 0.170 \\
MAE (mask $60\%$)  & 3.84              & 59.05                 & 149.60                & 40.84 & 50.0 & \underline{0.168} \\
MAE (mask $50\%$)  & \textbf{3.46}     & 57.48                 & \textbf{127.34}       & 31.02 & 52.1 & \textbf{0.164} \\
\bottomrule
\end{tabular*}
\end{table*}

\begin{figure*}[t]
\centering
\includegraphics[width=\linewidth]{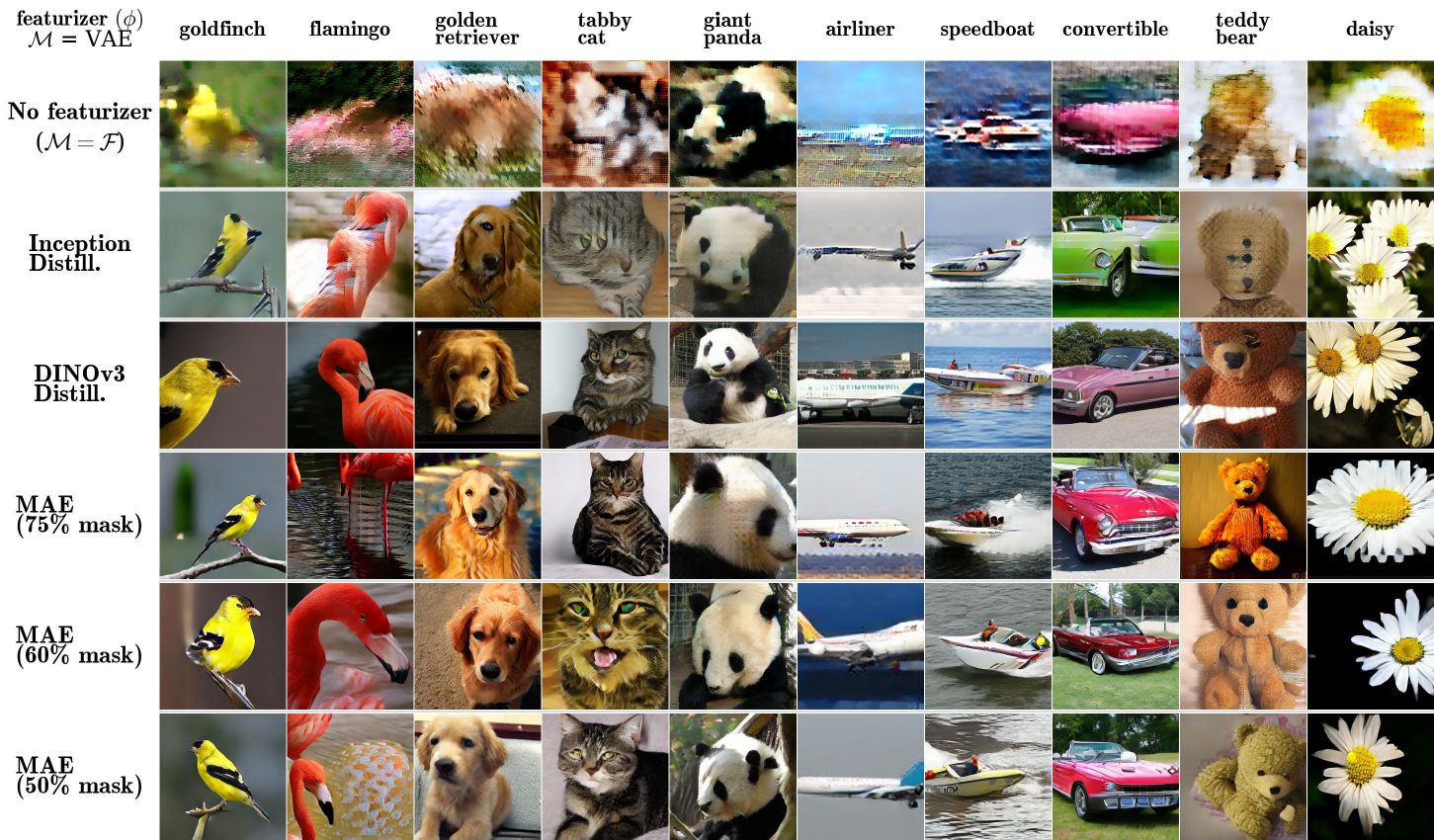}
\caption{%
  Uncurated class-conditional ImageNet samples from \semot{} (\Cref{sec:sinkhorn_method})
  (setup depicted in \Cref{fig:feature_geometry_setup}).
  Each row corresponds to a different frozen SSL feature extractor $\phi$ used during training. Note that the SSL feature extractor is not used at test time.
}
\label{fig:encoder_comparison}
\end{figure*}

\paragraph{The SSL featurizer is the primary lever.}
Without a featurizer, matching operates directly on VAE latent space ($\gM=\gF$): FID-Inception reaches $134.56$.
All SSL featurizers bring a dramatic improvement: FID-Inception drops to between $3.46$ and $6.66$ ($20$--$39\times$), and the gain holds on the independent FD-CLIP (from $195.85$ to $17.07$--$78.19$), reflecting the fact that VAE features fail to produce semantically coherent transport couplings (\Cref{fig:transport_plan}), while any SSL featurizer substantially sharpens class-level matching and sample quality (\Cref{fig:encoder_comparison}).
The choice of featurizer also matters: within the MAE family alone, FID-Inception varies from $6.66$ (mask $75\%$) to $3.46$ (mask $50\%$), a $1.9\times$ spread from a single hyperparameter, and cross-family differences ($3.46$ vs.\ $4.88$ for Inception distillation) are reflected in visible sample quality differences.

\paragraph{Geometric alignment alone does not determine generation quality.}
The alignment heatmap (\Cref{fig:encoder_alignment}) reveals the expected self-alignment pattern: each featurizer aligns best with its own evaluation counterpart ($R^2_{\mathrm{inc}}=0.46$ for Inception distillation, $R^2_{\mathrm{dino}}=0.20$ for DINOv3 distillation, $R^2_{\mathrm{vae}}=0.74$ for the no-featurizer baseline).
Yet across all three Fr\'echet metrics, the metric-matched featurizer is never the top performer: the best FID-Inception is MAE ($3.46$, not Inception distillation at $4.88$), the best FD-DINOv3 is MAE ($40.87$, not DINOv3 distillation at $43.87$), and the best FD-VAE is MAE ($127.34$, not the no-featurizer baseline at $179.80$).
On FD-CLIP, which matches no training featurizer, DINOv3 distillation ranks first ($17.07$) and MAE (mask $75\%$) second ($25.87$), the same order as under FD-DINOv2 (\Cref{app:independent_eval}), consistent with the DINOv3 teacher being itself a semantic encoder: the ranking among strong featurizers depends on the evaluation geometry, as the two terms of \eqref{eq:theory_main_bound} predict.
Matching estimation accounts for the advantage of MAE on the other metrics: the MAE family achieves lower effective rank ($52.1$--$57.3$ vs $68.0$ for DINOv3 and $87.0$ for Inception) and more stable transport targets ($\tilde{D}_{128}=0.164$--$0.170$ vs $0.191$ and $0.240$), making its Sinkhorn training signal far more reliable per minibatch.
Low effective rank alone is not sufficient, however: compressing the VAE latent with PCA lowers its effective rank by up to $40\times$ but leaves $\tilde{D}_{128}$ no better than the raw latent and its semantic alignment at zero (\Cref{app:pca_control}), so compactness helps only through semantic organization.
Matching stability thus emerges as a key driver of generation performance across featurizer families.
Single-cell RNA-seq experiments (\Cref{app:scrna}) further support this: within each generation space, $\tilde{D}_{128}$ consistently predicts mean Fr\'echet rank across evaluation spaces, suggesting that matching stability is an important criterion for identifying which feature extractors will lead to successful generation.

\paragraph{Metric hacking: the Inception case.}
The Inception-distilled featurizer achieves FID-Inception $= 4.88$, better than MAE (mask $75\%$) at $6.66$, yet \Cref{fig:encoder_comparison} shows its samples are substantially worse.
The explanation lies in geometric alignment: with $R^2_{\mathrm{inc}}=0.46$ against the FID evaluation featurizer (vs $0.12$ for MAE mask $75\%$), the Inception featurizer shapes the generated distribution toward Inception features, improving the reported metric without improving perceptual quality.
This is a concrete instance of metric hacking that FID alone cannot detect, in line with the concurrent finding that FID can misrank visual quality~\citep{yang2026representation}.
The independent evaluation confirms it: Inception distillation is the worst SSL featurizer on FD-CLIP ($78.19$ vs $17.07$--$40.84$ for the other SSL featurizers), as it is on FD-DINOv2, and it has the lowest precision and coverage under both encoders (\Cref{app:independent_eval}), so its FID-Inception score does not transfer to any encoder it did not target.
We further quantify this effect in \Cref{app:latent_fd}, where we re-evaluate using the distilled featurizers directly in VAE latent space as evaluation featurizers: the Inception-distilled model achieves FD-Inc-d $= 0.17$ while simultaneously reaching FD-VAE $= 509$, confirming that zero geometric misalignment relative to one evaluation space can mask catastrophically poor generation in any other.

\section{Conclusion}
\label{sec:conclusion}

One-step generative models trained by distribution matching are principled approaches to generation. With \semot{}, we show that lifting the matching objective to a semantically rich SSL feature space completely unlocks their potential, reducing FID-Inception by $39\times$ on class-conditional ImageNet. More importantly, we uncover the reasons: the mechanism is statistical, as semantic feature spaces have lower effective dimensionality, making matching estimation from finite minibatches reliable and matching stability a key criterion for feature extractor selection.
We also find that using the evaluation featurizer for training reduces geometric misalignment but can make matching estimation harder, improving the reported score while degrading sample quality. This metric hacking effect is particularly striking with Inception, the reference featurizer used by the community for FID, reinforcing the need for evaluation criteria that go beyond a single fixed metric.

Our study is limited in scale: we train a single DiT-B/2 generator on class-conditional ImageNet-256 with one autoencoder and our own featurizers, whereas Drifting~\citep{deng2026generative} shows that longer schedules and larger generators substantially improve one-step generation, a regime we did not explore.
More fundamentally, \semot{} relies on a feature space learned separately, for representation learning rather than for generation.
A natural end goal is to learn this space together with the generator, for instance by matching distributions in the model's own representation.
Our analysis indicates what such a space must provide, semantic alignment and stable minibatch transport, and what can go wrong: a feature space trained only to reduce the matching loss can collapse, so it needs an objective, such as a self-supervised one, that keeps it informative.
Turning our diagnostics into training signals for the feature space, and navigating the trade-off between representational richness and the geometric compactness that stable matching requires, are promising directions.

\begin{ack}
This work was funded by Genentech. All authors are employees of Genentech and hold equity in Roche.
\end{ack}

\bibliography{main}

\newpage

\appendix

\section*{Table of Contents}

\begin{itemize}[leftmargin=2em, itemsep=3pt, topsep=4pt]
  \item \textbf{\Cref{app:implementation}: Implementation Details.}
        Sinkhorn solvers and particle gradient, full \semot{} training pipeline, feature extraction, normalization, and hyperparameters.
  \item \textbf{\Cref{app:extra}: Additional Evaluations and Controls.}
        Latent-space evaluation with the distilled featurizers, which removes geometric misalignment by construction;
        Fr\'echet, kernel, fidelity, and coverage metrics in CLIP and DINOv2 spaces, which no training featurizer targets;
        and a PCA control showing that compact but non-semantic features are neither more stable nor semantically aligned.
  \item \textbf{\Cref{app:scrna}: Single-Cell RNA-Seq Generation.}
        Extended setup, diagnostics, and qualitative results for the single-cell domain.
  \item \textbf{\Cref{app:drifting_details}: Drifting and Sinkhorn Divergence: Detailed Comparison.}
        Precise relationship between the drifting objective and the Sinkhorn divergence gradient, including theoretical grounding and performance comparison.
  \item \textbf{\Cref{app:additional_samples}: Additional Generated Samples.}
        Uncurated class-conditional ImageNet samples for 30 further classes, for every featurizer.
\end{itemize}

\section{Implementation Details}
\label{app:implementation}

This section provides the Sinkhorn solvers and gradient used in training (\Cref{app:sinkhorn_details}), the training details and hyperparameters for the generator (\Cref{app:generator_training}), the SSL feature extractor training procedures (\Cref{app:ssl_training}), and the featurizer--evaluation alignment methodology (\Cref{app:encoder_alignment}). All experiments were run on NVIDIA B200 GPUs and are implemented on top of the \texttt{stable-pretraining-v1} framework~\citep{balestriero2025spt}, which provides the training loop, checkpointing, and configuration infrastructure.

\subsection{Sinkhorn Solvers and Gradient}
\label{app:sinkhorn_details}

This subsection provides the log-domain algorithms for computing $\mathrm{OT}_\varepsilon$ and the particle gradient of the Sinkhorn divergence used in training. Entropic OT and the Sinkhorn divergence are defined in \Cref{sec:background}; for full derivations see~\citep{cuturi2013sinkhorn,peyre2019computational,feydy2019interpolating}.

\paragraph{Sinkhorn iterations.}
The dual potentials $\vf, \vg \in \R^N$ solving $\mathrm{OT}_\varepsilon(\alpha, \beta)$ satisfy the fixed-point conditions
\begin{align}
\label{eq:app_sinkhorn_f}
    f_i &= \varepsilon \log(1/N) + \mine(C_{i\cdot} - \vg)\,, \\
\label{eq:app_sinkhorn_g}
    g_j &= \varepsilon \log(1/N) + \mine(C_{\cdot j} - \vf)\,,
\end{align}
where $\mine(\mathbf{S})_i \defeq -\varepsilon \log \sum_j \exp(-S_{ij}/\varepsilon)$ is the soft-minimum operator (converging to $\min_j S_{ij}$ as $\varepsilon\to 0$). Starting from $\vf = \vg = \mathbf{0}$, alternating these updates is the Sinkhorn algorithm~\citep{cuturi2013sinkhorn}. \Cref{alg:sinkhorn} gives the complete log-domain procedure.

\begin{algorithm}[h]
\caption{Sinkhorn algorithm (log-domain)}
\label{alg:sinkhorn}
\begin{algorithmic}[1]
\REQUIRE Cost matrix $\C \in \R^{N \times N}$, regularization $\varepsilon > 0$, iterations $K$
\STATE $f \gets \mathbf{0} \in \R^N$, \quad $g \gets \mathbf{0} \in \R^N$
\FOR{$\ell = 1, \dots, K$}
    \STATE $f_i \gets \varepsilon \log(1/N) + \mine(\C_{i\cdot} - \vg)$ for each $i$ \hfill \COMMENT{Update $\vf$}
    \STATE $g_j \gets \varepsilon \log(1/N) + \mine(\C_{\cdot j} - \vf)$ for each $j$ \hfill \COMMENT{Update $\vg$}
\ENDFOR
\STATE $\vpi \gets \tfrac{1}{N^2} \exp\!\big((\vf \oplus \vg - \C)/\varepsilon\big)$ \hfill \COMMENT{Recover optimal coupling}
\RETURN $\vpi$
\end{algorithmic}
\end{algorithm}

In our setting, $C_{ij} = \|\vh_i - \vh'_j\|^2$ for cross-transport and the returned coupling is $\vpi^{q,r}$; for self-transport use \Cref{alg:symmetric_sinkhorn}.

\paragraph{Symmetric Sinkhorn for self-transport.}
For the self-transport problem $\mathrm{OT}_\varepsilon(\alpha, \alpha)$, the dual problem is symmetric: $\vf = \vg$ at optimality. The optimal potential can therefore be found by a single \emph{symmetric} damped fixed-point update~\citep{feydy2019interpolating}:
\begin{equation*}
    f_i \;\leftarrow\; \tfrac{1}{2}\!\big(f_i + \varepsilon \log N + \mine(\C_{i\cdot} - \vf)\big)
    \qquad \text{for each } i,
\end{equation*}
which converges significantly faster than the standard alternating scheme: in practice $3$--$5$ iterations suffice. \Cref{alg:symmetric_sinkhorn} gives the complete procedure.

\begin{algorithm}[h]
\caption{Symmetric Sinkhorn algorithm (log-domain)}
\label{alg:symmetric_sinkhorn}
\begin{algorithmic}[1]
\REQUIRE Symmetric cost matrix $\C \in \R^{N \times N}$, regularization $\varepsilon > 0$, iterations $K$
\STATE $f \gets \mathbf{0} \in \R^N$
\FOR{$\ell = 1, \dots, K$}
    \STATE $f_i \gets \tfrac{1}{2}\!\big(f_i + \varepsilon \log N + \mine(\C_{i\cdot} - \vf)\big)$ for each $i$ \hfill \COMMENT{Symmetric update}
\ENDFOR
\STATE $\vpi \gets \tfrac{1}{N^2} \exp\!\big((\vf \oplus \vf - \C)/\varepsilon\big)$ \hfill \COMMENT{Recover symmetric coupling}
\RETURN $\vpi$
\end{algorithmic}
\end{algorithm}
In our setting, $C_{ij} = \|\vh_i - \vh_j\|^2$ for self-transport and the returned coupling is $\vpi^{q,q}$.

\paragraph{Particle gradient of $S_\varepsilon$.}
The gradient of $S_\varepsilon$ with respect to generated features recovers the entropic barycentric estimator of \citep{pooladian2024entropicestimationoptimaltransport}, linking the practical Sinkhorn update directly to the matching estimation theory of \Cref{sec:background}.
By the envelope theorem, treating the couplings as fixed at their optimal values~\citep{feydy2019interpolating}:
\begin{align}
    \nabla_{\vh_i} S_\varepsilon(\alpha,\beta)
    &= \nabla_{\vh_i}\mathrm{OT}_\varepsilon(\alpha,\beta)
       - \tfrac{1}{2}\,\nabla_{\vh_i}\mathrm{OT}_\varepsilon(\alpha,\alpha) \notag\\
    &= 2\!\sum_{j}\pi^{\alpha,\beta}_{ij}(\vh_i-\vh'_j)
       - \tfrac{1}{2}\!\Bigl[
           2\!\sum_k\pi^{\alpha,\alpha}_{ik}(\vh_i-\vh_k)
           + 2\!\sum_k\pi^{\alpha,\alpha}_{ki}(\vh_i-\vh_k)
         \Bigr] \notag\\
    &= 2\sum_{j} \pi^{\alpha,\beta}_{ij}(\vh_i - \vh'_j)
       \;-\; 2\sum_{k} \pi^{\alpha,\alpha}_{ik}(\vh_i - \vh_k)\,,
    \label{eq:sinkhorn_grad}
\end{align}
where $\vpi^{\alpha,\beta}$ and $\vpi^{\alpha,\alpha}$ are the optimal couplings for $\mathrm{OT}_\varepsilon(\alpha,\beta)$ and $\mathrm{OT}_\varepsilon(\alpha,\alpha)$ respectively ($\mathrm{OT}_\varepsilon(\beta,\beta)$ is free of $\{\vh_i\}$), and the last step uses $\pi^{\alpha,\alpha}_{ik}=\pi^{\alpha,\alpha}_{ki}$. Equivalently, in our setting, $\nabla_{\vh_i} S_\varepsilon = -\frac{2}{N}(\hat{T}_\varepsilon^{q,r}(\vh_i) - \hat{T}_\varepsilon^{q,q}(\vh_i))$, recovering \eqref{eq:theory_gradient_bridge}. The cross-transport term $\hat{T}_\varepsilon^{q,r}$ is the entropic barycentric estimator of \citep{pooladian2024entropicestimationoptimaltransport}: its convergence to the population Monge map at rate \eqref{eq:pnw_rate} is what makes the intrinsic dimensionality $d$ the key quantity controlling matching estimation quality per minibatch. The self-transport term $\hat{T}_\varepsilon^{q,q}$ is the debiasing correction inherited from the Sinkhorn divergence.

\subsection{Generator Training}
\label{app:generator_training}

This subsection details the training procedure for the one-step generator $\gen_\theta$.

\paragraph{Training loop.}
Each training step samples $N_c$ classes, generates $N_\text{neg}$ samples per class, draws $N_\text{pos}$ class-matched real targets and $N_\text{unc}$ unconditional real targets, and associates each class with a guidance weight $w_c$.
Each training step proceeds as follows:
\begin{enumerate}[leftmargin=2em, itemsep=1pt, topsep=2pt]
    \item Sample a batch of $N_c$ class labels and per-class guidance weights $w_c$.
    \item For each class $c$: generate $N_\text{neg}$ images, draw $N_\text{pos}$ positives and $N_\text{unc}$ unconditional samples from queues.
    \item Extract multi-scale features on all generated, positive, and unconditional samples.
    \item Normalize features; compute the conditional cross-, unconditional cross-, and self-transport targets; normalize.
    \item Compute the Sinkhorn divergence loss summed over all classes and features.
    \item Run backpropagation and update parameters; update EMA.
\end{enumerate}
The remainder of this section elaborates on each component.

\paragraph{Data loading, queues, and batching.}
At each training step, $N_c$ class labels are sampled uniformly at random and $N_\text{neg}$ images are generated per class, yielding an effective batch size $B = N_c \times N_\text{neg}$. Real samples are provided by a standard DataLoader and used only to populate two CPU-resident queues~\citep{he2020momentum}: a \emph{per-class} queue of size $Q_c = 128$ for positive samples and a \emph{global} unconditional queue of size $Q_u = 1000$ for CFG negatives. At each step, $N_\text{pos}$ positive samples are drawn from the class-matched queue and $N_\text{unc}$ unconditional samples from the global queue.

\paragraph{Multi-scale feature extraction via attention keys.}
\label{app:key_facet}
To capture feature structure at multiple levels of abstraction, we extract features from
several intermediate layers of the ViT backbone rather than relying on the final block
output alone.
At each tapped layer we use \emph{attention key vectors}~\citep{amir2021deep} rather than
block output tokens: key vectors are computed before self-attention aggregation and MLP
mixing, and therefore retain finer spatial structure than the corresponding block outputs.

For the backbone described in \Cref{app:ssl_training} (depth~$32$), we tap every $8$ layers (layers $8,16,24,32$).
At each tapped layer, we apply \texttt{norm1} (the pre-attention layer norm), project through the QKV linear layer,
select the key slice, apply \texttt{k\_norm} (key normalization from QK-norm, if present), and reshape into a spatial map
of shape $(B, C, H_p, W_p)$, where $H_p = W_p = 16$ for $32{\times}32$ SD-VAE latents
with patch size~$2$ and $C = 1280$ (see \Cref{app:ssl_training}).
From each spatial map we form three feature blocks:
\begin{itemize}[leftmargin=2em, itemsep=1pt, topsep=2pt]
  \item \textbf{per\_loc}: the $H_pW_p = 256$ spatial key vectors, shape $(B, 256, C)$;
  \item \textbf{global\_mean}: their mean-pooled vector, shape $(B, 1, C)$;
  \item \textbf{global\_std}: their std-pooled vector, shape $(B, 1, C)$.
\end{itemize}
This yields $4 \times 3 = 12$ named blocks.
Each spatial position within a block defines an independent OT sub-problem~\citep{deng2026generative}: for a \textbf{per\_loc} block with $256$ positions, there are $256$ independent Sinkhorn problems, each matching $N$-sample empirical measures in $\R^C$; the two global blocks each contribute one sub-problem in $\R^C$.
In total, a ViT-H encoder produces $4 \times (256 + 1 + 1) + 1 = 1033$ independent OT sub-problems per class (the final $+1$ is the input-norm feature: the per-channel spatial RMS of the input VAE latent, a vector in $\R^4$ that contributes one additional OT sub-problem).
Per-location matching provides spatially resolved gradient signal: each position independently constrains the generator to produce correct local content at that grid location.
Scalar featurizer diagnostics (effective rank, $\tilde{D}_N$) are aggregated across
sub-problems by a weighted average, with weight equal to the number of spatial positions in the
block ($256$ for \textbf{per\_loc}, $1$ for global blocks), so the reported value is
dominated by the per-location sub-problems.

Each feature type $\phi_j$ is normalized independently so that pairwise distances are of order~$1$ regardless of feature dimension, ensuring that the Sinkhorn regularization $\varepsilon$ controls the entropy--fidelity tradeoff consistently across all sub-problems.
The normalization scale is
\begin{equation*}
    S_j = \frac{\mathrm{scale}_j}{\sqrt{C_j}}\,,
\end{equation*}
where $C_j = C$ is the feature dimension and $\mathrm{scale}_j$ is computed with stop-gradient as the empirical mean $L_2$ distance from generated features to the full target pool $[\phi_j(\vx_{\mathrm{gen}}), \phi_j(\vx_{\mathrm{unc}}), \phi_j(\vx_{\mathrm{pos}})]$, concatenated across all spatial positions within the block~\citep{deng2026generative}. The Sinkhorn regularization strength is then set data-adaptively: $\tilde{\varepsilon} = \tau \cdot \mathrm{std}(\mathbf{C}^{q,r})$, where $\mathrm{std}(\mathbf{C}^{q,r})$ is the empirical standard deviation of the cross-transport cost matrix entries (computed after feature normalization), making the base temperatures $\tau \in \{0.02, 0.05, 0.2\}$ scale-invariant.

\paragraph{OT barycenter normalization.}
After computing $\hat{T}_\varepsilon^{\mathrm{cfg}}$ (defined in~\eqref{eq:cfg_field}) for each $(\phi_j, \varepsilon)$ pair, the normalization scale is
\begin{equation*}
    \lambda_j = \sqrt{\E\!\Big[\tfrac{1}{C_j}\bigl\|\hat{T}_\varepsilon^{\mathrm{cfg}}(\vh_i)\bigr\|^2\Big]}\,,
\end{equation*}
and the normalized transport target is $\tilde{T}_j = \hat{T}_\varepsilon^{\mathrm{cfg}} / \lambda_j$. We use three regularization values $\varepsilon \in \{0.02, 0.05, 0.2\}$: for each~$\varepsilon$, $\tilde{T}_{j,\varepsilon}$ is computed and normalized independently, and the aggregated target is $\tilde{T}_j = \sum_\varepsilon \tilde{T}_{j,\varepsilon}$. All normalization statistics are computed with stop-gradient. Features at different spatial locations within the same feature map share a single normalization scale, computed by concatenating all locations.

\paragraph{Classifier-free guidance.}
CFG is implemented via the Sinkhorn CFG objective~\eqref{eq:cfg_objective}. For each class~$c$, three Sinkhorn problems are solved: the conditional cross-transport between $q_{\phi,\theta}(\cdot|c,w_c)$ and $r_\phi^c$, the unconditional cross-transport between $q_{\phi,\theta}(\cdot|c,w_c)$ and $r_\phi^u$, and the self-transport of $q_{\phi,\theta}(\cdot|c,w_c)$. They are combined into $\hat{T}_\varepsilon^{\mathrm{cfg}}$~\eqref{eq:cfg_field} with guidance weight $w_c \ge 0$. Following~\citep{deng2026generative}, the self-transport cost matrix has its diagonal set to $+\infty$ ($C^{q,q}_{ii} = +\infty$, see~\Cref{app:drifting_details}) to prevent trivial self-matching, and is solved via the symmetric Sinkhorn algorithm (\Cref{alg:symmetric_sinkhorn}). The self-transport solve is shared across both Sinkhorn divergences. The $N_\text{unc}$ unconditional real samples drawn per class from the global queue serve as the target for the unconditional cross-transport solve. At each training step, $w_c$ is sampled from the power-law distribution $p(w) \propto w^{-\kappa}$ on $[w_{\min}, w_{\max}]$ via inverse-CDF sampling, and the generator is conditioned on $(c, w_c)$ via adaptive layer normalization (adaLN)~\citep{peebles2023scalable}. At inference, a single forward pass with a user-specified~$w$ produces the output; no Sinkhorn computation is needed at test time.

\paragraph{Optimizer and EMA.}
The total loss $\mathcal{L} = \sum_j \mathcal{L}_j$ sums over feature types after averaging each term over its class-location entries. We use AdamW~\citep{loshchilov2017decoupled} with $(\beta_1, \beta_2) = (0.9, 0.95)$, gradient clipping at norm~$2.0$, and a linear warmup followed by linear decay schedule. An EMA copy of the generator with decay~$0.999$ is updated after each step.

\subsection{SSL Feature Extractor Training}
\label{app:ssl_training}

This subsection details the training procedures for the SSL feature extractors used to define the matching cost during generator training.

\paragraph{Featurizer architecture.}
All featurizers share the same transformer backbone: hidden dimension~$1280$, depth~$32$, $16$ attention heads, SwiGLU MLP (hidden dim~$5120$), RMSNorm, and RMSNorm applied to queries and keys before attention (QK-norm), operating on SD-VAE latents of shape $32\times32\times4$ (scaled by $0.18215$) with patch size~$2$, yielding $256$ tokens.
This common capacity and input format ensures that differences in generation quality and OT diagnostics across featurizer families reflect differences in SSL training objective rather than model size or input representation.

\paragraph{Training setup.}
All featurizers are trained with AdamW ($\beta_1=0.9$, $\beta_2=0.95$, weight decay~$0.05$) in bf16 mixed precision, with an EMA copy (decay~$0.9995$) used for checkpointing.
Input images undergo pixel-space augmentation (RandomResizedCrop at $256\times256$, RandomHorizontalFlip) followed by on-the-fly SD-VAE encoding with posterior sampling.

We now describe the training objective specific to each featurizer family.

\subsubsection{MAE Featurizer Pretraining}
\label{app:mae_pretrain}
Each MAE feature extractor is pretrained from scratch with masked autoencoding directly on SD-VAE latents.

\paragraph{Decoder architecture.}
The decoder uses hidden dimension~$512$, depth~$4$, and $16$ heads.

\paragraph{Pretraining.}
Masking operates at the token level (after patch embedding): a fraction $m$ of the $256$ tokens is randomly removed before the encoder ($m\in\{50\%,\,60\%,\,75\%\}$), and the decoder predicts the original latent patches for all masked positions.
The loss is MSE restricted to masked positions.
Training uses learning rate $1.5\times10^{-4}$ with $10\text{k}$-step linear warmup and global batch size~$4096$, running for $1280$ epochs (${\approx}400\text{k}$ steps).

\paragraph{Classification fine-tuning.}
For all three mask variants, we append $3000$ additional steps with a combined loss $\mathcal{L} = (1-\lambda)\mathcal{L}_{\mathrm{recon}} + \lambda\mathcal{L}_{\mathrm{cls}}$, where $\lambda$ is linearly warmed from $0$ to $0.1$ over the first $1000$ steps.
$\mathcal{L}_{\mathrm{cls}}$ is the cross-entropy loss of a linear head ($1280\to1000$) applied to the encoder CLS token.
This step adds a weak semantic signal without discarding the reconstruction geometry.

\subsubsection{DINOv3 Distillation}
\label{app:dinov3_distill}
The DINOv3-distilled featurizer trains a student with the architecture described above on SD-VAE latents to match the features of a frozen DINOv3 ViT-7B teacher operating on pixels.

\paragraph{Distillation.}
Two linear projection heads map student features to the teacher dimension: a global head ($1280\to4096$) applied to the CLS token and a patch head ($1280\to4096$) applied to all $256$ patch tokens.
The teacher is a frozen DINOv3 ViT-7B-16 pretrained on pixels~\citep{simeoni2025dinov3}; it receives $256\times256$ pixel crops (scale $[0.4, 1.0]$, to avoid tiny crops that degrade patch-token supervision) and produces a $4096$-dim CLS token and $256$ patch tokens.
The student receives the corresponding SD-VAE latents and produces its own CLS and patch tokens.
Both sides are L2-normalized before computing cosine similarity losses on CLS ($\lambda_{\mathrm{cls}}=0.1$) and patch tokens ($\lambda_{\mathrm{patch}}=1.0$).
Training uses learning rate $10^{-4}$ with $5\text{k}$-step linear warmup and global batch size~$2048$, for $200$ epochs (${\approx}125\text{k}$ steps).

\subsubsection{Inception Distillation}
\label{app:inception_distill}
The Inception-distilled featurizer trains a student with the architecture described above on SD-VAE latents to match the pool3 features of a frozen InceptionV3 network (the same model used for FID computation~\citep{heusel2017gans}).

\paragraph{Distillation.}
A single linear projection head ($1280\to2048$) maps the mean-pooled student tokens to the Inception feature space.
The teacher is a frozen InceptionV3~\citep{szegedy2016rethinking}, receiving $256\times256$ pixel crops (scale $[0.4, 1.0]$, converted to uint8 $[0,255]$) and producing $2048$-dim pool3 features.
The student receives the corresponding SD-VAE latents and produces a global descriptor by mean-pooling its $256$ patch tokens.
Both sides are L2-normalized before computing the cosine similarity loss between the projected student features and the teacher pool3 features.
Training uses the same schedule as DINOv3 distillation: learning rate $10^{-4}$ with $5\text{k}$-step linear warmup and global batch size~$2048$, for $200$ epochs (${\approx}125\text{k}$ steps).

\subsection{Featurizer--Evaluation Alignment}
\label{app:encoder_alignment}

The geometric misalignment term $\Delta^\mathrm{geom}_\phi$ in the decomposition \eqref{eq:theory_main_bound} is the key quantity linking the training featurizer $\phi$ to the evaluation metric: when $\phi$ and $\phi_\star$ induce similar matching geometries, improving the Sinkhorn training objective translates into gains on the evaluation metric; when they diverge, the two objectives partially decouple.
Computing $\Delta^\mathrm{geom}_\phi(\gen_\theta) = J_\star(\gen_\theta) - J_\phi(\gen_\theta)$ directly requires a trained generator, making it a post-hoc diagnostic rather than a tool for featurizer selection.
We instead measure a featurizer-level proxy: the held-out $R^2$ of a linear map $A\colon\gF\to\gF_\star$ fitting $A\phi(\vx)\approx\phi_\star(\vx)$, which is computable before any generation training and captures whether $\phi_\star$'s geometry is linearly recoverable from $\phi$.

The $R^2$ is fitted via kernel ridge regression on 2048 ImageNet-256 latents ($\lambda=0.1$, features L2-normalised per sample before fitting).
$R^2 \approx 1$ means $\phi$ already contains $\phi_\star$'s information linearly (small geometric misalignment); $R^2 \approx 0$ means the two featurizers live in unrelated spaces.
Three evaluation featurizers are used: InceptionV3 pool3 (2048-dim), DINOv3 ViT-7B CLS token
(4096-dim), and the SD-VAE flat latent scaled by $0.18215$ (4096-dim).
The resulting $R^2$ heatmap for the six ViT-H featurizers and its interpretation are in
\Cref{fig:encoder_alignment} and the surrounding discussion in \Cref{sec:exp_imagenet}.

\section{Additional Evaluations and Controls}
\label{app:extra}

This section collects three complementary analyses of the featurizers in \Cref{tab:imagenet_results}: evaluation in the distilled latent feature spaces, which removes geometric misalignment by construction (\Cref{app:latent_fd}); evaluation with independent encoders and diversity metrics (\Cref{app:independent_eval}); and a PCA control separating compactness from semantic structure (\Cref{app:pca_control}).

\subsection{Latent-Space Evaluation with Distilled Featurizers}
\label{app:latent_fd}

\Cref{tab:imagenet_results} evaluates in pixel-space feature spaces (InceptionV3, DINOv3 ViT-7B) that differ from the distilled latent-space featurizers used during training.
To isolate the effect of geometric alignment, \Cref{tab:latent_fd} evaluates the same generators using these distilled latent-space featurizers directly as $\phi_\star$.
By construction, this sets $\Delta^\mathrm{geom}_\phi = 0$ (\Cref{eq:theory_main_bound}), leaving matching estimation as the sole driver of the training-evaluation gap and providing a controlled test of its relative importance.

When the training featurizer matches the evaluation featurizer, FD drops sharply: Inception-distilled training achieves FD-Inc-d $=0.17$, and DINOv3-distilled training achieves FD-DINOv3-d $=3.03$, both far below any cross-featurizer evaluation.
However, the Inception-distilled model that scores $0.17$ in its own feature space simultaneously achieves FD-VAE $=509$, the worst latent-space reconstruction of any featurizer, confirming that near-perfect self-alignment is an artifact of matching the evaluation geometry rather than a sign of generation quality.
The MAE $50\%$ featurizer achieves the best FD-MAE50 ($1.76$) and, notably, the best FD-Inc-d ($0.08$), outperforming the Inception-distilled featurizer even in the latter's own matched evaluation space, confirming that matching estimation quality is an independent driver of generation performance, strong enough to compensate for geometric misalignment.

\begin{table}[ht]
\centering
\caption{%
  Latent-space Fr\'echet distance: evaluation with distilled featurizers applied directly to VAE latents.
  Each column uses a different latent-space featurizer as $\phi_\star$.
  \colorbox{blue!8}{Shading}: training featurizer~$\phi$ matches the evaluation featurizer.
}
\label{tab:latent_fd}
\small
\begin{tabular}{lcccc}
\toprule
Featurizer & FD-VAE$\downarrow$ & FD-DINOv3-d$\downarrow$ & FD-Inc-d$\downarrow$ & FD-MAE50$\downarrow$ \\
\midrule
No featurizer ($\gM=\gF$) & 179.93 & 156.59 & 4.49  & 125.58 \\
Inception distill.         & 508.77 & 13.91  & \cellcolor{blue!8}0.17  & 10.31 \\
DINOv3 distill.            & \textbf{100.26} & \cellcolor{blue!8}\textbf{3.03}  & \underline{0.09}  & \underline{1.74} \\
MAE (mask $75\%$)          & \underline{126.01} & 5.80   & 0.17  & 2.68  \\
MAE (mask $50\%$)          & 126.82 & \underline{5.45}  & \textbf{0.08}  & \cellcolor{blue!8}\textbf{1.76} \\
\bottomrule
\end{tabular}
\end{table}

\subsection{Independent Evaluation Encoders and Diversity Metrics}
\label{app:independent_eval}

Two evaluation featurizers of \Cref{tab:imagenet_results} are related to training featurizers: FID-Inception and FD-DINOv3 apply the teachers of the Inception- and DINOv3-distilled featurizers, although in pixel space and with a different architecture than the latent-space students used for training.
To remove this overlap, \Cref{tab:independent_eval} evaluates the same six generators, at the same operating points, with two encoders involved neither in training nor in distillation: CLIP ViT-L/14~\citep{radford2021clip} (pooled image embedding, 768-dim) and DINOv2 ViT-L~\citep{oquab2024dinov2} (CLS token, 1024-dim), the backbone recommended for FD evaluation~\citep{stein2023exposing}. DINOv2 belongs to the same model family as the DINOv3 teacher, so CLIP is the strictly independent check.
Besides FD, we report KID~\citep{binkowski2018demystifying} (unbiased MMD$^2$ with a cubic polynomial kernel, averaged over 100 subsets of 1000 samples), precision and recall~\citep{kynkaanniemi2019improved}, and density and coverage~\citep{naeem2020reliable} ($k=5$ nearest neighbors on 10k samples).
Precision and density measure fidelity; recall and coverage measure diversity.
All metrics compare 50k generated samples with the 50k ImageNet validation images, resized to $224\times224$.

Both conclusions of \Cref{sec:experiments} hold under these encoders.
First, every SSL featurizer improves on the no-featurizer baseline by a wide margin. Without a featurizer, recall is $0.00$ and coverage $0.01$ under CLIP, i.e.\ near-total mode collapse; its high precision and density under DINOv2 reflect samples concentrated on few modes, which both fidelity metrics reward, and are not a sign of quality.
Second, Inception distillation is the weakest SSL featurizer on every metric of both encoders except DINOv2 density, confirming that its FID-Inception score (\Cref{tab:imagenet_results}) does not reflect sample quality in spaces it was not trained to match.
Among SSL featurizers, DINOv3 distillation is best on every CLIP metric and on FD, KID, recall, and coverage under DINOv2, and the gains are not obtained by trading coverage for fidelity.

\begin{table}[ht]
\centering
\caption{%
  Independent-encoder evaluation of the generators of \Cref{tab:imagenet_results}.
  KID is reported $\times 10^3$.
  $\downarrow$: lower is better; $\uparrow$: higher is better.
  \textbf{Bold}: best SSL featurizer per column (the no-featurizer baseline is excluded because mode collapse inflates its precision and density).
}
\label{tab:independent_eval}
\small
\begin{tabular}{lcccccc}
\toprule
Featurizer & FD$\downarrow$ & KID$\downarrow$ & Prec.$\uparrow$ & Rec.$\uparrow$ & Dens.$\uparrow$ & Cov.$\uparrow$ \\
\midrule
\multicolumn{7}{l}{\textit{CLIP ViT-L/14}} \\
No featurizer ($\gM=\gF$) & 195.85 & 755.1 & 0.67 & 0.00 & 0.37 & 0.01 \\
Inception distill.         & 78.19  & 290.8 & 0.48 & 0.21 & 0.28 & 0.37 \\
DINOv3 distill.            & \textbf{17.07} & \textbf{43.9} & \textbf{0.82} & \textbf{0.64} & \textbf{0.76} & \textbf{0.84} \\
MAE (mask $75\%$)          & 25.87  & 74.3  & 0.80 & 0.42 & 0.74 & 0.77 \\
MAE (mask $60\%$)          & 40.84  & 130.3 & 0.71 & 0.39 & 0.56 & 0.65 \\
MAE (mask $50\%$)          & 31.02  & 91.8  & 0.78 & 0.44 & 0.68 & 0.72 \\
\midrule
\multicolumn{7}{l}{\textit{DINOv2 ViT-L}} \\
No featurizer ($\gM=\gF$) & 1657.73 & 3725.4 & 0.98 & 0.00 & 8.37 & 0.11 \\
Inception distill.         & 409.50  & 554.0  & 0.79 & 0.57 & \textbf{1.00} & 0.54 \\
DINOv3 distill.            & \textbf{78.25} & \textbf{57.7} & \textbf{0.90} & \textbf{0.88} & 0.89 & \textbf{0.91} \\
MAE (mask $75\%$)          & 138.69  & 107.6  & \textbf{0.90} & 0.73 & 0.95 & 0.87 \\
MAE (mask $60\%$)          & 221.87  & 230.9  & 0.85 & 0.72 & 0.85 & 0.74 \\
MAE (mask $50\%$)          & 189.38  & 190.2  & 0.85 & 0.78 & 0.81 & 0.77 \\
\bottomrule
\end{tabular}
\end{table}

\subsection{Compactness without Semantics: a PCA Control}
\label{app:pca_control}

Among SSL featurizers, a lower effective rank goes with more stable transport targets (\Cref{tab:imagenet_results}).
To test whether compactness alone explains this, we compress the raw VAE latent with PCA, which lowers its dimension without adding semantic structure, and measure both gaps of~\eqref{eq:theory_main_bound} at the featurizer level.
We project flattened SD-VAE latents of ImageNet-256 onto their top $k\in\{16,32,64,128,256\}$ principal components, and compute $\tilde{D}_{128}$ as for \Cref{tab:imagenet_results} (held-out pool of 2048 latents, six replicates) and the alignment $R^2$ as in \Cref{app:encoder_alignment}.
No generator is trained on these features: the test is at the featurizer level, and its consequence for generation is a prediction of~\eqref{eq:theory_main_bound} rather than a measurement.

\Cref{tab:pca_control} shows that PCA lowers the effective rank of the latent by $6$--$40\times$, yet $\tilde{D}_{128}$ does not improve: it stays at $0.220$--$0.240$, slightly above the raw latent ($0.206$; $0.194$ in \Cref{tab:imagenet_results}, within the replicate spread) and well above the MAE featurizers ($0.164$--$0.170$) and DINOv3 distillation ($0.191$).
Its alignment with the semantic evaluation featurizers is zero, as for the raw latent, while it stays partly aligned with the VAE geometry it is derived from ($R^2_{\mathrm{vae}}=0.31$--$0.47$).
Compression alone therefore lowers neither gap: PCA keeps the highest-variance directions of the VAE latent, which, as the zero alignment shows, carry little semantic information.
Effective rank tracks matching stability among SSL featurizers because there a low rank results from suppressing nuisance detail; it is a diagnostic rather than an independent lever.
The single-cell experiments (\Cref{app:scrna}) give the converse case: there, PCA is among the best matching spaces, because its top-variance directions are the cell-type-relevant ones.

\begin{table}[ht]
\centering
\caption{%
  PCA control on SD-VAE latents.
  Effective rank is computed on the 2048-sample pool, so its values are not comparable to \Cref{tab:imagenet_results}.
  $\tilde{D}_{128}$: mean $\pm$ std over six replicates.
  $R^2$: held-out alignment as in \Cref{fig:encoder_alignment}; negative values are shown as $\approx 0$.
  SSL reference rows report $\tilde{D}_{128}$ from \Cref{tab:imagenet_results} and $R^2$ from \Cref{fig:encoder_alignment}; MAE row: range over the three mask ratios.
}
\label{tab:pca_control}
\small
\begin{tabular}{lcccccc}
\toprule
Features & Dim & Eff.\ rank & $\tilde{D}_{128}{\downarrow}$ & $R^2_{\mathrm{inc}}$ & $R^2_{\mathrm{dino}}$ & $R^2_{\mathrm{vae}}$ \\
\midrule
Raw VAE latent & 4096 & 394.7 & $0.206 \pm 0.016$ & $\approx 0$ & $\approx 0$ & 0.74 \\
PCA-VAE        & 16   & 9.9   & $0.240 \pm 0.045$ & $\approx 0$ & $\approx 0$ & 0.31 \\
PCA-VAE        & 32   & 16.0  & $0.233 \pm 0.036$ & $\approx 0$ & $\approx 0$ & 0.37 \\
PCA-VAE        & 64   & 25.2  & $0.234 \pm 0.033$ & $\approx 0$ & $\approx 0$ & 0.41 \\
PCA-VAE        & 128  & 40.6  & $0.227 \pm 0.028$ & $\approx 0$ & $\approx 0$ & 0.44 \\
PCA-VAE        & 256  & 70.0  & $0.220 \pm 0.024$ & $\approx 0$ & $\approx 0$ & 0.47 \\
\midrule
\multicolumn{7}{l}{\textit{SSL featurizers (reference)}} \\
MAE (mask $50$--$75\%$) & -- & -- & $0.164$--$0.170$ & $0.09$--$0.12$ & $0.04$--$0.06$ & $0.27$--$0.28$ \\
DINOv3 distill.         & -- & -- & $0.191$ & 0.21 & 0.20 & 0.22 \\
\bottomrule
\end{tabular}
\end{table}

\section{Single-Cell RNA-Seq Generation}
\label{app:scrna}

\paragraph{Setup.}
We apply \semot{} to cell-type-conditional single-cell RNA-seq generation on a standard scRNA-seq stimulation dataset~\citep{kang2018multiplexed} ($16{,}839$ cell profiles, $3{,}000$ genes after standard preprocessing).
The generator $\gen_\theta$ is a 6-layer MLP mapping Gaussian noise to the generation space~$\gM$, conditioned on cell-type labels via one-hot concatenation.
Three representation spaces are considered:
\emph{ambient} ($\R^{3000}$, the log-normalized gene expression vector),
\emph{PCA} ($\R^{128}$, the top principal components of the ambient space, a commonly used single cell representation in generative models~\citep{rohbeck2025modeling,andersson2026single}), and
\emph{scVI} ($\R^{30}$, the latent embedding of a pretrained variational autoencoder for single-cell data~\citep{lopez2018deep}).
We compare six configurations that vary the generation space~$\gM$ and the Sinkhorn coupling space~$\phi$:
\textbf{PCA\,/\,scVI} ($\gM = \R^{128}_{\mathrm{PCA}}$, $\phi = \mathrm{scVI}$),
\textbf{ambient\,/\,scVI} ($\gM = \R^{3000}$, $\phi = \mathrm{scVI}$),
\textbf{PCA\,/\,PCA} ($\gM = \R^{128}_{\mathrm{PCA}}$, $\phi = \mathrm{Id}$),
\textbf{ambient\,/\,ambient} ($\gM = \R^{3000}$, $\phi = \mathrm{Id}$),
\textbf{ambient\,/\,PCA} ($\gM = \R^{3000}$, $\phi = \mathrm{PCA}$),
and \textbf{scVI\,/\,scVI} ($\gM = \R^{30}_{\mathrm{scVI}}$, $\phi = \mathrm{Id}$).
Generation quality is measured by conditional and unconditional Fr\'echet distances in three evaluation spaces: the 30-dimensional scVI latent, the 128-dimensional PCA space, and the 3000-dimensional ambient expression space.
Each configuration is run over 5 seeds.

\begin{figure}[t]
\centering
\includegraphics[width=\linewidth]{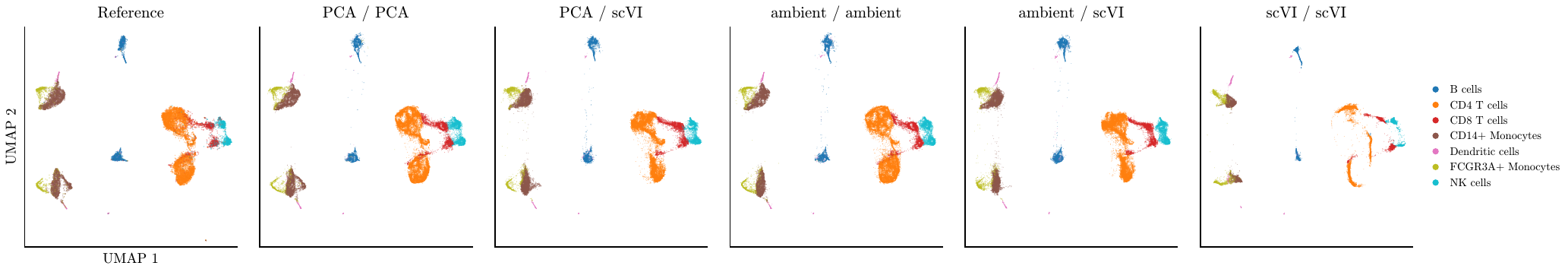}
\caption{%
  UMAP visualization of reference and generated cells for six configurations, colored by cell type.
  PCA\,/\,PCA and PCA\,/\,scVI closely reproduce the reference cluster geometry, while scVI\,/\,scVI collapses several populations.%
}
\label{app:scrna_umap}
\end{figure}

\paragraph{Results.}
\Cref{app:scrna_results} reports generation quality alongside coupling-space diagnostics, and \Cref{app:scrna_rank_summary} aggregates these into a mean rank across all twelve Fr\'echet distance metrics.
As in the image setting, the OT stability metric $\tilde{D}_{128}$ is predictive of generation quality.
Within a given generation space, the coupling with lowest $\tilde{D}_{128}$ consistently achieves the best mean FD rank.
Among PCA-space generators, PCA coupling ($\tilde{D}_{128}=0.54$) ranks first ($1.67$) while scVI coupling ($\tilde{D}_{128}=0.89$) ranks $2.83$.
Among ambient-space generators, PCA coupling again leads (rank $3.00$), ahead of ambient ($\tilde{D}_{128}=0.91$, rank $3.67$) and scVI coupling (rank $3.83$).
Generating directly in the scVI latent (scVI\,/\,scVI) is consistently worst, as the 30-dimensional generation space is too restrictive for the generator to express a sufficiently rich transport map.
\Cref{app:scrna_umap} confirms this visually: PCA\,/\,PCA and PCA\,/\,scVI closely reproduce the reference cluster geometry, whereas scVI\,/\,scVI collapses several cell-type populations.

\begin{table}[t]
\centering
\caption{%
  Single-cell generation results (Kang dataset, cell-type-conditional, test set).
  Fr\'echet distances computed in three evaluation spaces.
  Effective rank and $\tilde{D}_{128}$ characterize the coupling embedding space.
  Mean $\pm$ SEM over 5 seeds.
  \textbf{Bold}: best per column; \underline{underline}: second best;
  \colorbox{blue!8}{shading}: training featurizer~$\phi$ matches the evaluation featurizer (diagonal entries).%
}
\label{app:scrna_results}
\resizebox{\textwidth}{!}{%
\begin{tabular}{lcccccccc}
\toprule
Config ($\gM$ / $\phi$) & Cond.\ FD$_{\text{scVI}}{\downarrow}$ & FD$_{\text{scVI}}{\downarrow}$ & Cond.\ FD$_{\text{PCA}}{\downarrow}$ & FD$_{\text{PCA}}{\downarrow}$ & Cond.\ FD$_{\text{amb}}{\downarrow}$ & FD$_{\text{amb}}{\downarrow}$ & Eff.\ rank & $\tilde{D}_{128}{\downarrow}$ \\
\midrule
PCA / PCA         & $2.28 \pm 0.01$             & $1.26 \pm 0.02$             & \cellcolor{blue!8}$\bm{22.84} \pm 0.22$        & \cellcolor{blue!8}$\bm{15.71} \pm 0.03$        & $\bm{93.16} \pm 0.22$        & $\bm{78.08} \pm 0.03$        & 20.0  & $\bm{0.54}$ \\
PCA / scVI        & \cellcolor{blue!8}$\bm{1.77} \pm 0.04$        & \cellcolor{blue!8}$\bm{0.78} \pm 0.03$        & $24.91 \pm 0.15$             & $\underline{17.41} \pm 0.08$ & $95.20 \pm 0.15$             & $79.69 \pm 0.07$             & 16.3  & $\underline{0.89}$ \\
ambient / PCA     & $2.51 \pm 0.02$             & $1.50 \pm 0.01$             & \cellcolor{blue!8}$\underline{24.43} \pm 0.44$ & \cellcolor{blue!8}$17.80 \pm 0.03$             & $\underline{94.13} \pm 0.45$ & $78.64 \pm 0.05$             & 20.0  & $\bm{0.54}$ \\
ambient / ambient & $2.55 \pm 0.03$             & $1.51 \pm 0.02$             & $24.90 \pm 0.58$             & $17.82 \pm 0.04$             & \cellcolor{blue!8}$94.62 \pm 0.58$             & \cellcolor{blue!8}$\underline{78.60} \pm 0.04$ & 260.8 & $0.91$ \\
ambient / scVI    & \cellcolor{blue!8}$\underline{1.81} \pm 0.02$ & \cellcolor{blue!8}$\underline{0.90} \pm 0.03$ & $26.72 \pm 0.24$             & $19.37 \pm 0.09$             & $96.05 \pm 0.24$             & $79.58 \pm 0.08$             & 16.3  & $\underline{0.89}$ \\
scVI / scVI       & \cellcolor{blue!8}$3.02 \pm 0.02$             & \cellcolor{blue!8}$2.24 \pm 0.01$             & $50.47 \pm 0.09$             & $36.62 \pm 0.05$             & $124.21 \pm 0.11$            & $101.14 \pm 0.06$            & 16.3  & $\underline{0.89}$ \\
\bottomrule
\end{tabular}%
}
\end{table}

\begin{table}[t]
\centering
\caption{%
  Mean rank across all twelve Fr\'echet distance metrics in \Cref{app:scrna_results}
  (3 evaluation spaces $\times$ conditional/unconditional $\times$ val/test).
  Lower $\tilde{D}_{128}$ of the coupling space consistently predicts lower (better) mean FD rank
  within a given generation space, mirroring the ImageNet finding (\Cref{sec:experiments}).%
}
\label{app:scrna_rank_summary}
\begin{tabular}{lccc}
\toprule
Config ($\gM$ / $\phi$) & $\tilde{D}_{128}{\downarrow}$ & Eff.\ rank & Mean FD rank${\downarrow}$ \\
\midrule
PCA / PCA & \textbf{0.54} & 20.0 & \textbf{1.67} \\
PCA / scVI & 0.89 & 16.3 & 2.83 \\
\addlinespace
ambient / PCA & \textbf{0.54} & 20.0 & \textbf{3.00} \\
ambient / ambient & 0.91 & 260.8 & 3.67 \\
ambient / scVI & 0.89 & 16.3 & 3.83 \\
\addlinespace
scVI / scVI & 0.89 & 16.3 & 6.00 \\
\bottomrule
\end{tabular}
\end{table}

\paragraph{Conclusion.} \Cref{app:scrna_rank_summary} shows that OT stability is predictive of which feature space leads to best generation in general. PCA shows the lowest $\tilde{D}_{128}$ and boasts the best performance across the different model spaces (PCA and ambient). While the effective rank appears less predictive of performance in this case, the difference between PCA (20) and scVI (16.3) is marginal. Finally, we note that while scVI is a viable choice of feature space, using it as a model space severely degrades performance, which we attribute to the poor performance of the scVI decoder. This highlights an important strength of our approach: leveraging a model space where reconstruction is highly accurate (but OT stability may be low) and a feature space where OT stability is high (but reconstruction may be poor).

\section{Drifting and Sinkhorn Divergence: Detailed Comparison}
\label{app:drifting_details}

Drifting~\citep{deng2026generative} is a recent one-step generative model that, like \semot{}, uses frozen SSL features to define a transport objective in latent space. This appendix spells out the drifting construction and explains its precise relationship to the Sinkhorn divergence.

\paragraph{Construction.}
Let $\vh_i = \phi(\vu_i)$ for $i=1,\ldots,N$ be feature vectors of $N$ generated samples and $\vh_j^r = \phi(\vu_j^r)$ for $j=1,\ldots,M$ be feature vectors of $M$ real samples.
Define the pairwise squared-Euclidean costs
\begin{equation}
\label{eq:app_costs}
    C^{q,r}_{ij} = \|\vh_i - \vh_j^r\|^2\,,
    \qquad
    C^{q,q}_{ik} = \|\vh_i - \vh_k\|^2\,,
    \qquad
    C^{q,q}_{ii} = +\infty\,,
\end{equation}
where the diagonal mask eliminates self-interactions. These are concatenated column-wise into a single cost matrix $\bar{C} \in \R^{N \times (M+N)}$, pooling real and generated samples as joint targets.

Drifting then applies a symmetric row/column softmax normalization on $\bar{C}$ at temperature $\tau > 0$:
\begin{equation}
\label{eq:app_double_softmax}
    A_{it}
    \defeq
    \sqrt{
        \frac{\exp(-\bar{C}_{it}/\tau)}{\sum_s \exp(-\bar{C}_{is}/\tau)}
        \cdot
        \frac{\exp(-\bar{C}_{it}/\tau)}{\sum_\ell \exp(-\bar{C}_{\ell t}/\tau)}
    }\,,
\end{equation}
so $A_{it}$ is the geometric mean of a row-wise and a column-wise softmax: a weight is large only when the pair is cheap both within its source row and within its target column.

Splitting $A$ into real and generated blocks $A^+ \in \R^{N\times M}$ and $A^- \in \R^{N\times N}$, the rows of each block are rescaled so that each particle $i$ contributes equal total mass to attraction and repulsion:
\[
    W^+_{ij} = \frac{A^+_{ij}}{\sum_j A^+_{ij}}, \qquad W^-_{ik} = \frac{A^-_{ik}}{\sum_k A^-_{ik}}.
\]
The resulting drifting field is
\[
    \mathbf{V}_i = \sum_j W^+_{ij}\,\vh_j^r - \sum_{k \neq i} W^-_{ik}\,\vh_k,
\]
and the model is trained by regressing $\vh_i$ toward $\mathrm{sg}[\vh_i + \mathbf{V}_i]$, where $\mathrm{sg}[\cdot]$ denotes stop-gradient.

\paragraph{Relation to Sinkhorn divergence.}
The double-softmax in \eqref{eq:app_double_softmax} is built from the same Gibbs kernel $\exp(-C/\tau)$ as entropic OT: each row softmax normalizes over source particles and each column softmax over target particles, so $A_{it}$ approximates a coupling density. Splitting $A$ into a cross block $A^+$ (generated vs.\ real) and a self block $A^-$ (generated vs.\ generated) then mirrors the attraction--repulsion structure of the Sinkhorn divergence gradient: $A^+$ attracts generated particles toward real data and $A^-$ pushes them apart, replicating the cross- and self-transport terms of $\nabla S_\varepsilon$.

Two key differences separate drifting from the Sinkhorn divergence gradient:
\begin{itemize}[leftmargin=2em, itemsep=4pt, topsep=4pt]
  \item \textbf{One step vs.\ convergence.}
        The double-softmax in \eqref{eq:app_double_softmax} amounts to a single iteration of
        Sinkhorn-Knopp: one round of row normalization and one of column normalization applied
        simultaneously. A single step does not produce a doubly stochastic coupling: $A$
        satisfies neither its prescribed row marginals nor its column marginals, so mass is not
        properly allocated to each source or target particle. \semot{} iterates two independent
        Sinkhorn solves until both marginal constraints are jointly satisfied, yielding properly
        normalized couplings (\Cref{app:sinkhorn_details}).
  \item \textbf{Coupled vs.\ decoupled transports.}
        Drifting normalizes a single concatenated cost matrix, so the attraction block $A^+$ and
        the repulsion block $A^-$ share the same marginal budget. \semot{} solves two entirely
        separate problems (one for $q_{\phi,\theta}$ against $r_\phi$ and one for $q_{\phi,\theta}$ against itself),
        so attraction and repulsion carry independent marginal constraints and cannot interfere
        with each other.
  \item \textbf{Theoretical grounding.}
        The Sinkhorn divergence is a theoretically principled objective: it interpolates between
        the Wasserstein distance and the MMD~\citep{feydy2019interpolating}, and its population
        limit is directly linked to the $W_2$ distance that underlies Fr\'echet-style evaluation
        metrics. This connection is what motivates using it as a training objective for generation
        quality measured by FID and related scores (\Cref{sec:eval_metrics}). The drifting
        objective, by contrast, lacks this theoretical interpretation: the double-softmax
        normalization and cross-scaling do not correspond to the gradient of any known
        population-level divergence, making it difficult to reason about what the model
        optimizes at convergence.
\end{itemize}

\paragraph{Performance comparison and scaling.}
\semot{} with a DiT-B/2 generator achieves FID-Inception $3.46$
(\Cref{tab:imagenet_results}), close to the $3.36$ reported by \citep{deng2026generative}
at the same generator scale.
Direct comparison is limited, however, for two reasons.
First, our SSL featurizers use a ViT backbone trained with our own recipes
(\Cref{app:mae_pretrain,app:dinov3_distill,app:inception_distill}),
while drifting uses a ResNet featurizer; the backbone architecture and training procedure
materially shape the transport geometry and generation quality, making the feature
extractors non-interchangeable.
Second, as the metric hacking analysis in \Cref{sec:experiments} shows, FID is sensitive
to the choice of training featurizer: a featurizer well-aligned to Inception features can
improve the reported score while degrading actual sample quality, so FID alone is an
unreliable basis for cross-method comparison when featurizers differ.
Scaling the generator from DiT-B/2 to DiT-L/2, as drifting demonstrates, remains a
direct path toward competitive generation performance.

\section{Additional Generated Samples}
\label{app:additional_samples}

We provide additional uncurated class-conditional ImageNet samples across 30 diverse classes, complementing the 10 classes shown in \Cref{fig:encoder_comparison}.
Each figure follows the same layout: rows correspond to different frozen SSL feature extractors used during training, and columns correspond to ImageNet classes.
All samples are generated in one step and decoded by the frozen SD-VAE decoder; the SSL feature extractor is not used at inference.

\begin{figure*}[ht]
\centering
\includegraphics[width=\linewidth]{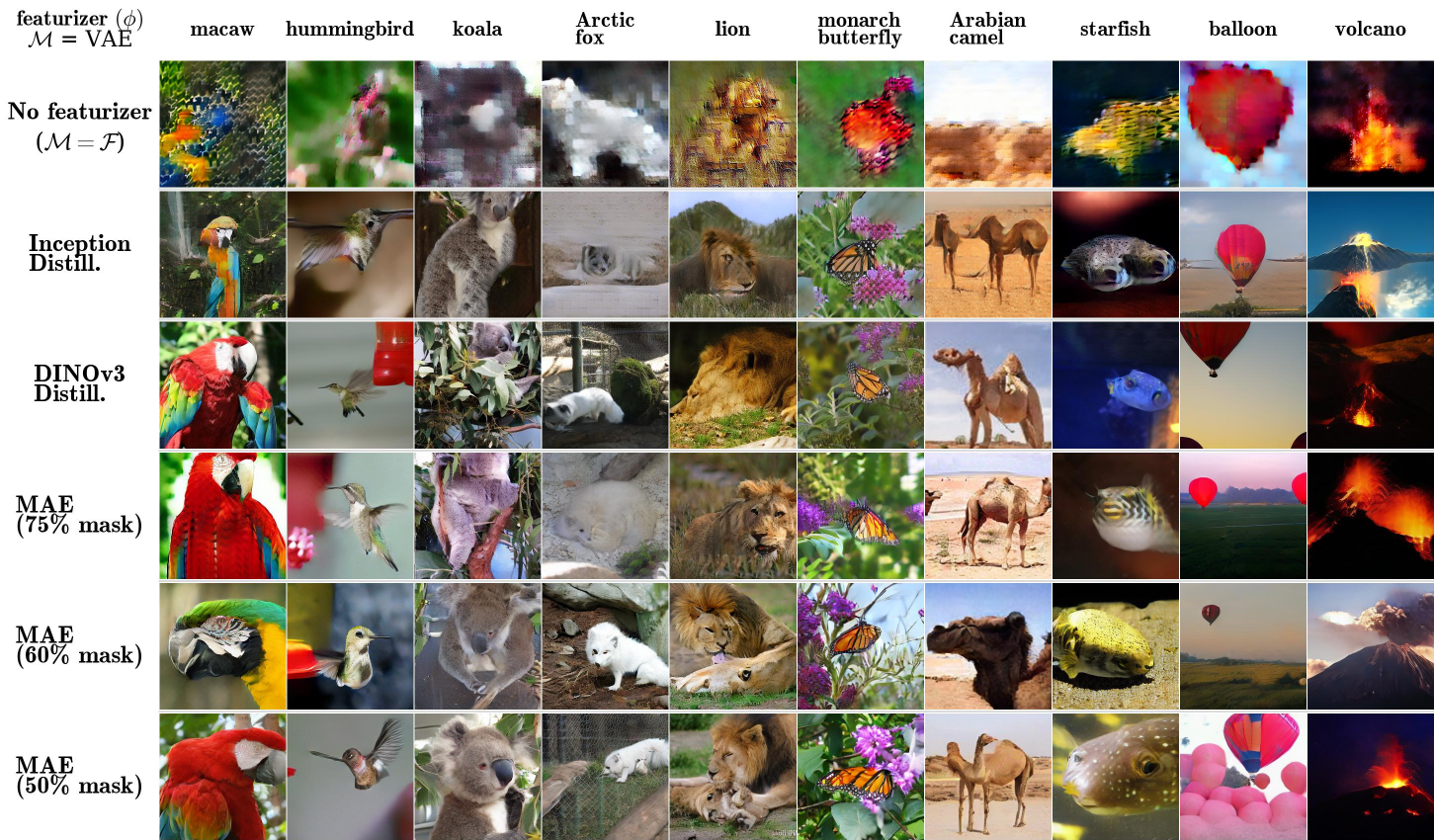}
\caption{%
  Additional uncurated samples (classes: macaw, hummingbird, koala, Arctic fox, lion, monarch butterfly, Arabian camel, starfish, balloon, volcano).
  Same setup as \Cref{fig:encoder_comparison}.%
}
\label{fig:encoder_comparison_panelB}
\end{figure*}

\begin{figure*}[ht]
\centering
\includegraphics[width=\linewidth]{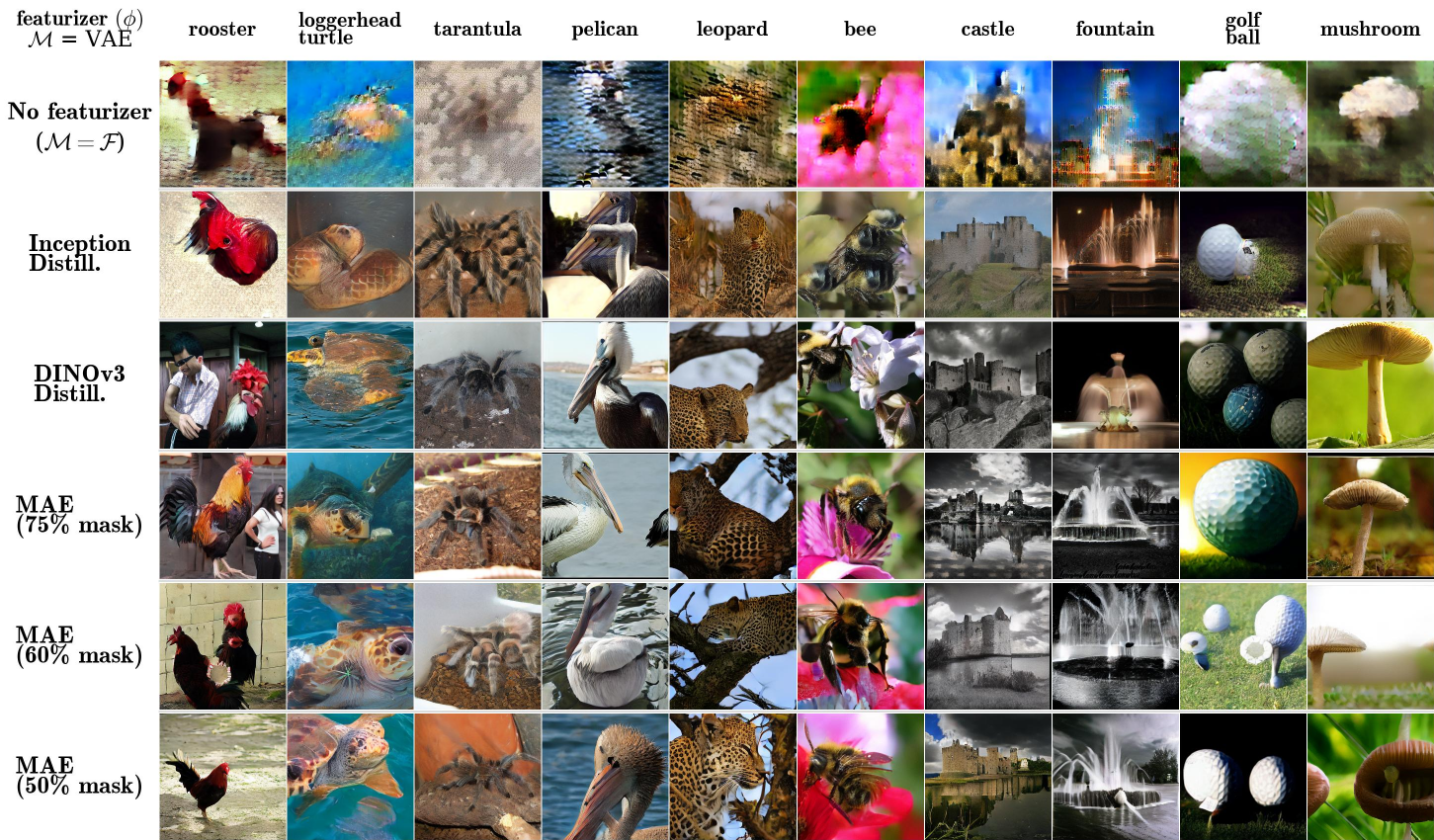}
\caption{%
  Additional uncurated samples (classes: rooster, loggerhead turtle, tarantula, pelican, leopard, bee, castle, fountain, golf ball, mushroom).
  Same setup as \Cref{fig:encoder_comparison}.%
}
\label{fig:encoder_comparison_panelC}
\end{figure*}

\begin{figure*}[ht]
\centering
\includegraphics[width=\linewidth]{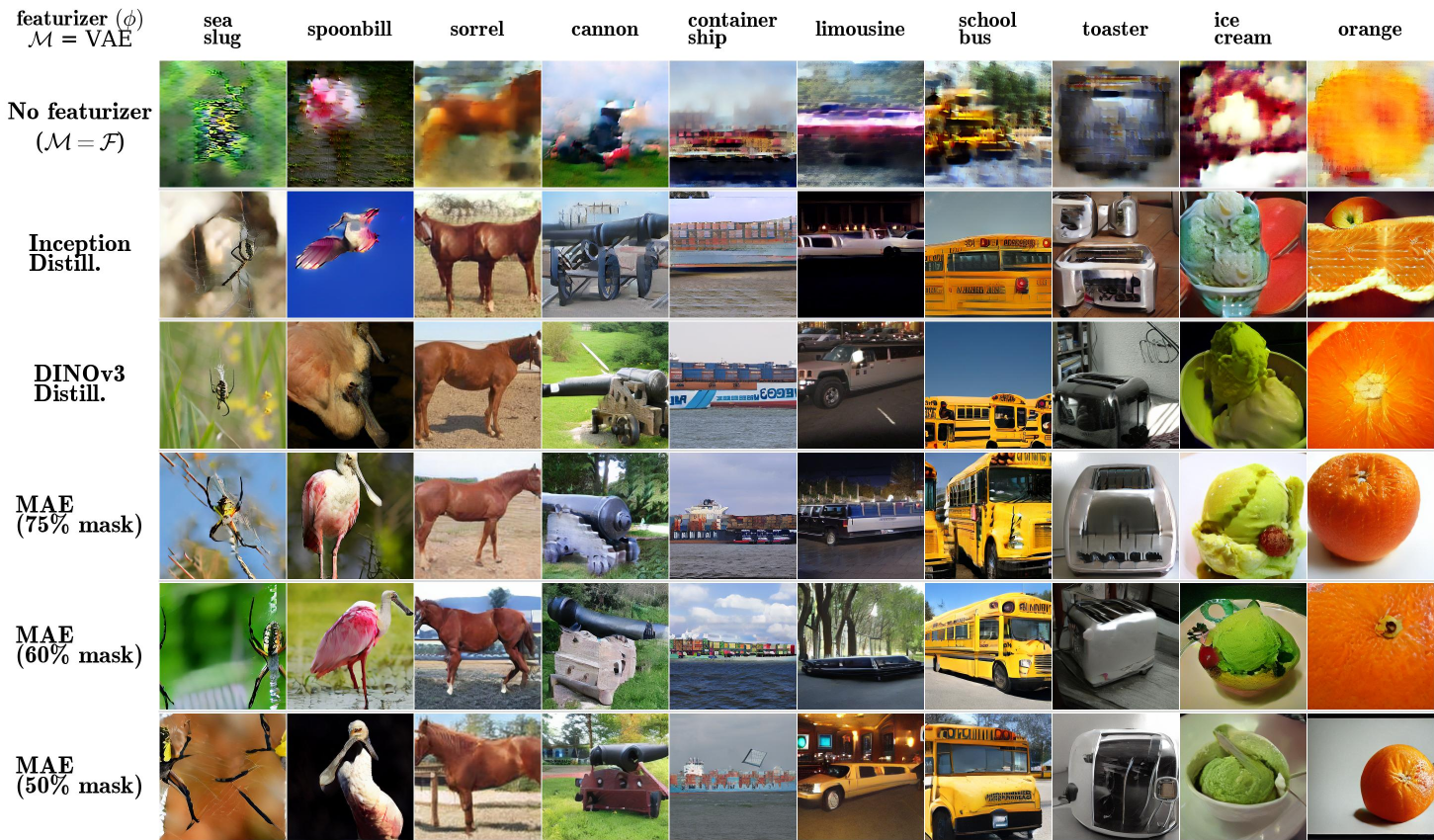}
\caption{%
  Additional uncurated samples (classes: sea slug, spoonbill, sorrel, cannon, container ship, limousine, school bus, toaster, ice cream, orange).
  Same setup as \Cref{fig:encoder_comparison}.%
}
\label{fig:encoder_comparison_panelD}
\end{figure*}

\end{document}